\documentclass[sigconf,nonacm]{acmart}
\usepackage{algorithm}  
\usepackage{algpseudocode}
\usepackage{subfigure}
\usepackage{multirow}
\usepackage{multicol}
\usepackage{pbox}

\AtBeginDocument{%
  \providecommand\BibTeX{{%
    \normalfont B\kern-0.5em{\scshape i\kern-0.25em b}\kern-0.8em\TeX}}}

\copyrightyear{2020} 
\acmYear{2020} 
\setcopyright{acmcopyright}\acmConference[KDD '20]{Proceedings of the 26th ACM SIGKDD Conference on Knowledge Discovery and Data Mining}{August 23--27, 2020}{Virtual Event, USA}
\acmBooktitle{Proceedings of the 26th ACM SIGKDD Conference on Knowledge Discovery and Data Mining (KDD '20), August 23--27, 2020, Virtual Event, USA}
\acmPrice{15.00}
\acmDOI{10.1145/3394486.3403125}
\acmISBN{978-1-4503-7998-4/20/08}

\begin{document}
\fancyhead{}
\title{TIPRDC: Task-Independent Privacy-Respecting Data Crowdsourcing Framework for Deep Learning with Anonymized Intermediate Representations}

\author{Ang Li$^1$, Yixiao Duan$^2$, Huanrui Yang$^1$, Yiran Chen$^1$, Jianlei Yang$^2$}
\affiliation{
\institution{$^1$Department of Electrical and Computer Engineering, Duke University\\
$^2$School of Computer Science and Engineering, Beihang University}
$^1$\{ang.li630, huanrui.yang, yiran.chen\}@duke.edu, $^2$\{jamesdyx, jianlei\}@buaa.edu.cn}

\begin{abstract}
The success of deep learning partially benefits from the availability of various large-scale datasets. These datasets are often crowdsourced from individual users and contain private information like gender, age, etc. 
The emerging privacy concerns from users on data sharing hinder the generation or use of crowdsourcing datasets and lead to hunger of training data for new deep learning applications. One na\"{\i}ve solution is to pre-process the raw data  to extract features at the user-side, and then only the extracted features will be sent to the data collector. Unfortunately, attackers can still exploit these extracted features to train an adversary classifier to infer private attributes. Some prior arts leveraged game theory to protect private attributes. 
However, these defenses are designed for known primary learning tasks, the extracted features work poorly for unknown learning tasks. 
To tackle the case where the learning task may be unknown or changing, we present \textit{TIPRDC}, a task-independent privacy-respecting data crowdsourcing framework with anonymized intermediate representation. 
The goal of this framework is to learn a feature extractor that can hide the privacy information from the intermediate representations; while maximally retaining the original information embedded in the raw data for the data collector to accomplish unknown learning tasks.
We design a hybrid training method to learn the anonymized intermediate representation: (1) an adversarial training process for \textit{hiding private information from features}; (2) \textit{maximally retain original information} using a neural-network-based mutual information estimator. We extensively evaluate TIPRDC and compare it with existing methods using two image datasets and one text dataset. Our results show that TIPRDC substantially outperforms other existing methods. Our work is the first task-independent privacy-respecting data crowdsourcing framework.
\end{abstract}

\ccsdesc[500]{Security and privacy}
\ccsdesc[500]{Computing methodologies~Machine learning}

\keywords{privacy-respecting data crowdsourcing; anonymized intermediate representations; deep learning}

\maketitle

\section{Introduction}\label{sec:introduction}
Deep learning has demonstrated an impressive performance in many applications, such as computer vision \cite{krizhevsky2012imagenet,he2016deep} and natural language processing \cite{bahdanau2014neural,wu2016google,oord2016wavenet}. Such success of deep learning partially benefits from various large-scale datasets (e.g., ImageNet \cite{deng2009imagenet}, MS-COCO \cite{lin2014microsoft}, etc.), which can be used to train powerful deep neural networks (DNN). The datasets are often crowdsourced from individual users to train DNN models. For example, companies or research institutes that want to implement face recognition systems may collect the facial images from employees or volunteers. However, those data that are crowdsourced from individual users for deep learning applications often contain private information such as gender, age, etc. Unfortunately, the data crowdsourcing process can be exposed to serious privacy risks as the data may be misused by the data collector or acquired by the adversary.
It is recently reported that many large companies face data security and user privacy challenges. The data breach of Facebook, for example, raises users' severe concerns on sharing their personal data. These emerging privacy concerns hinder generation or use of large-scale crowdsourcing datasets and lead to hunger of training data of many new deep learning applications. 
A number of countries are also establishing laws to protect data security and privacy. 
As a famous example, the new European Union‘s General Data Protection Regulation (GDPR) requires companies to not store personal data for a long time, and allows users to delete or withdraw their personal data within 30 days. It is critical to design a data crowdsourcing framework to protect the privacy of the shared data while maintaining the utility for training DNN models.
\begin{figure*}[t]
    \centering
    \includegraphics[scale=0.3]{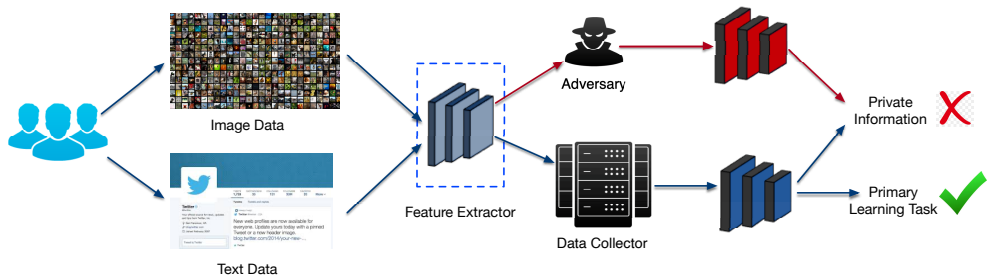}\vspace{-0.5cm}
    \caption{The overview of TIPRDC.}
    \label{fig:overview}
\end{figure*}

Existing solutions to protect privacy are struggling to balance the tradeoff between privacy and utility. 
An obvious and widely adopted solution is to transform the raw data into task-oriented features, and users only upload the extracted features to corresponding service providers, such as Google Now \cite{googlenow} and Google Cloud \cite{googlecloud}. 
Even though transmitting only features are generally more secure than uploading raw data, recent developments in model inversion attacks \cite{mahendran2015understanding,dosovitskiy2016inverting,dosovitskiy2016generating} have demonstrated that adversaries can exploit the acquired features to reconstruct the raw image, and hence the person on the raw image can be re-identified from the reconstructed image. 
In addition, the extracted features can also be exploited by an adversary to infer private attributes, such as gender, age, etc. Ossia \textit{et al.} \cite{osia2020hybrid} move forward by applying dimentionality reduction and noise injection to the features before uploading them to the service provider. However, such approach leads to unignorable utility loss. Inspired by Generative Adversarial Networks (GAN), several adversarial learning approaches \cite{liu2019privacy,li2019deepobfuscator,oh2017adversarial,kim2019training} have been proposed to learn obfuscated features from raw images. Unfortunately, those solutions are designed for known primary learning tasks, which limits their applicability in the data crowdsourcing where the primary learning task may be unknown or changed when training a DNN model. 
The need of collecting large-scale crowdsourcing dataset under strict requirement of data privacy and limited applicability of existing solutions motivates us to design a privacy-respecting data crowdsourcing framework: the raw data from the users are locally transformed into an intermediate representation that can remove the private information while retaining the discriminative features for primary learning tasks.

In this work, we propose TIPRDC -- a task-independent privacy-respecting data crowdsourcing framework with anonymized intermediate representation. The ultimate goal of this framework is to learn a feature extractor that can remove the privacy information from the extracted intermediate features while maximally retaining the original information embedded in the raw data for primary learning tasks. 
As Figure \ref{fig:overview} shows, participants can locally run the feature extractor and submit only those intermediate representations to the data collector instead of submitting the raw data. 
The data collector then trains DNN models using these collected intermediate representations, but both the data collector and the adversary cannot accurately infer any protected private information. 
Compared with existing adversarial learning methods \cite{liu2019privacy,li2019deepobfuscator,oh2017adversarial,kim2019training}, TIPRDC does not require the knowledge of the primary learning task and hence, directly applying existing adversarial training methods becomes impractical. 
It is challenging to remove all concerned private information that needs to be protected while retaining everything else for unknown primary learning tasks. 
To address this issue, we design a hybrid learning method to learn the anonymized intermediate representation. The learning purpose is two-folded: (1) hiding private information from features; (2) maximally retaining original information. Specifically, we hide private information from features by performing our proposed privacy adversarial training (PAT) algorithm, which simulates the game between an adversary who makes efforts to infer private attributes from the extracted features and a defender who aims to protect user privacy. The original information are retained by applying our proposed MaxMI algorithm, which aims to maximize the mutual information between the feature of the raw data and the union of the private information and the retained feature.

In summary, our key contributions are the follows:
\begin{itemize}
    \item To the best of our knowledge, TIPRDC is the first privacy-respecting data crowdsourcing framework for deep learning without the knowledge of any specific primary learning task. By applying TIPRDC, the learned feature extractor can hide private information from features while maximally retaining the information of the raw data.
    \item We propose a privacy adversarial training algorithm to enable the feature extractor to hide privacy information from features. In addition, we also design the MaxMI algorithm to maximize the mutual information between the raw data and the union of the private information and the retained feature, so that the original information from the raw data can be maximally retained in the feature.
    \item We quantitatively evaluate the utility-privacy tradeoff with applying TIPRDC on three real-world datasets, including both image and text data. We also compare the performance of TIPRDC with existing solutions.
\end{itemize}

The rest of this paper is organized as follows. Section \ref{sec:related_work} reviews the related work. Section \ref{sec:problem_formulation} presents the problem formulation. Section \ref{sec:framework_design} describes the framework overview and details of core modules. Section \ref{sec:evaluation} evaluates the framework.  Section \ref{sec:conclusion} concludes this paper.
\section{Related Work}\label{sec:related_work}
\textbf{Data Privacy Protection}: Many techniques have been proposed to protect data privacy, most of which are based on various anonymization methods including \textit{k}-anonymity \cite{sweeney2002k}, \textit{l}-diversity \cite{mahendran2015understanding} and \textit{t}-closeness \cite{li2007t}. 
However, these approaches are designed for protecting sensitive attributes in a static database and hence, are not suitable to our addressed problem -- data privacy protection in the online data crowdsourcing for training DNN models. 
Differential privacy \cite{duchi2013local,erlingsson2014rappor,bassily2015local,qin2016heavy,smith2017interaction,avent2017blender,wang2017locally} is another widely applied technique to protect privacy of an individual's data record, which provides a strong privacy guarantee. However, the privacy guarantee provided by differential privacy is different from the privacy protection offered by TIPRDC in data crowdsourcing. The goal of differential privacy is to add random noise to a user’s true data record such that two arbitrary true data records have close probabilities to generate the same noisy data record. 
Compared with differential privacy, our goal is to hide private information from the features such that an adversary cannot accurately infer the protected private information through training DNN models. Osia \textit{et al.} \cite{osia2020hybrid} leverage a combination of dimensionality reduction, noise addition, and Siamese fine-tuning to protect sensitive information from features, but it does not offer the tradeoff between privacy and utility in a systematic way.

\textbf{Visual Privacy Protection:} Some works have been done to specifically preserve privacy in images and videos. De-identification is a typical privacy-preserving visual recognition approach to alter the raw image such that the identity cannot be visually recognized. 
There are various techniques to achieve de-identification, such as Gaussian blur \cite{oh2016faceless}, identity obfuscation \cite{oh2016faceless}, mean shift filtering \cite{winkler2014trusteye} and adversarial image perturbation \cite{oh2017adversarial}. 
Although those approaches are effective in protecting visual privacy, they all limit the utility of the data for training DNN models. 
In addition, encryption-based approaches \cite{gilad2016cryptonets,yonetani2017privacy} have been proposed to guarantee the privacy of the data, but they require specialized DNN models to directly train on the encrypted data. Unfortunately, such encryption-based solutions prevent general dataset release and introduce substantial computational overhead. All the above practices only consider protecting privacy in specific data format, i.e., image and video, which limit their applicability across diverse data modalities in the real world.

\textbf{Tradeoff between Privacy and Utility using Adversarial Networks}: With recent advances in deep learning, several approaches have been proposed to protect data privacy using adversarial networks and simulate the game between the attacker and the defender who defend each other with conflicting utility-privacy goals. 
Pittaluga \textit{et al.} \cite{pittaluga2019learning} design an adversarial learning method for learning an encoding function to defend against performing inference for specific attributes from the encoded features. 
Seong \textit{et al.} \cite{oh2017adversarial} introduce an adversarial network to obfuscate the raw image so that the attacker cannot successfully perform image recognition. Wu \textit{et al.} \cite{wu2018towards} design an adversarial framework to explicitly learn a degradation transform for the original video inputs, aiming to balance between target task performance and the associated privacy budgets on the degraded video. 
Li \textit{et al.} \cite{li2019deepobfuscator} and Liu \textit{et al.} \cite{liu2019privacy} propose approaches to learn obfuscated features using adversarial networks, and only obfuscated features will be submitted to the service provider for performing inference. Attacker cannot train an adversary classifier using collected obfuscated features to accurately infer a user's private attributes or reconstruct the raw data.
The same idea behind the above solutions is that using adversarial networks to obfuscate the raw data or features, in order to defending against privacy leakage. However, those solutions are designed to protect privacy information while targeting some specified learning tasks, such as face recognition, activity recognition, etc. 
Our proposed TIPRDC provides a more general privacy protection, which does not require  the knowledge of the primary learning task.

\textbf{Differences between TIPRDC and existing methods:} Compared with prior arts, our proposed TIPRDC has two distinguished features: (1) a more general approach that be applied on different data formats instead of handling only image data or static database; (2) require no knowledge of primary learning tasks.
\section{Problem Formulation}\label{sec:problem_formulation}
There are three parties involved in the crowdsourcing process: \textit{user}, \textit{adversary}, and \textit{data collector}. 
Under the strict requirement of data privacy, a data collector offers options to a user to specify any private attribute that needs to be protected. Here we denote the private attribute specified by a user as $u$. 
According to the requirement of protecting $u$, the data collector will learn a feature extractor  $f_\theta(z | x,u)$ that is parameterized by weight $\theta$, which is the core of TIPRDC. 
The data collector distributes the data collecting request associated with the feature extractor to users. Given the raw data $x$ provided by a user, the feature extractor can locally extract feature $z$ from $x$ while hiding private attribute $u$. 
Then, only extracted feature $z$ will be shared with the data collector, which can training DNN models for primary learning tasks using collected $z$. An adversary, who may be an authorized internal staff of the data collector or an external hacker, has access to the extracted feature $z$ and aims to infer private attribute $u$ based on $z$. We assume an adversary can train a DNN model via collecting $z$, and then the trained model takes a user's extracted feature $z$ as input and infers the user's private attribute $u$.

The critical challenge of TIPRDC is to learn the feature extractor, which can hide private attribute from features while maximally retaining original information from the raw data. 
Note that it is very likely that the data and private attribute are somehow correlated. Therefore, we cannot guarantee no information about $u$ will be contained in $z$ unless we enforce it in the objective function when training the feature extractor.

The ultimate goal of the feature extractor $f$ is two-folded:
\begin{itemize}
    \item \textbf{Goal 1:} make sure the extracted features conveys no private attribute;
    \item \textbf{Goal 2:} retain as much information of the raw data as possible to maintain the utility for primary learning tasks.
\end{itemize}

\begin{figure}[tb]
\centering
    \captionsetup{width=1.0\linewidth}
		\includegraphics[width=0.6\linewidth]{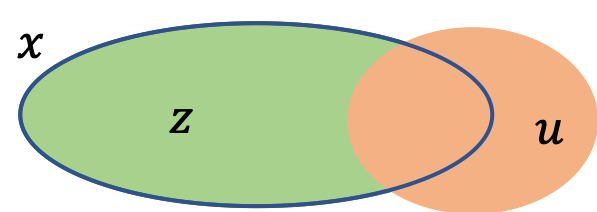}
	\caption{Optimal outcome of feature extractor $f_\theta(z|x,u)$.}
	\label{fig:set}
\end{figure}

Formally, Goal 1 can be formulated as:
\begin{equation}
    \label{equ:min}
    \min_\theta I(z; u),
\end{equation}
where $I(z; u)$ represents the mutual information between $z$ and $u$. On the other hand, Goal 2 can be formulated as:
\begin{equation}
    \label{equ:max}
    \max_\theta I(x; z | u).
\end{equation}
In order to avoid any potential conflict with the objective of Goal 1, we need to mitigate counting in the information where $u$ and $x$ are correlated. Therefore, the mutual information in Equation~\ref{equ:max} is evaluated under the condition of private attribute $u$. Note that
\begin{equation}
    I(x; z | u) = I(x; z,u)-I(x;u).
\end{equation}
Since both $x$ and $u$ are considered fixed under our setting, $I(x;u)$ will stay as a constant during the optimization process of feature extractor $f_\theta$. Therefore, we can safely rewrite the objective of Goal 2 as:
\begin{equation}
    \label{equ:altmax}
    \max_\theta I(x; z,u),
\end{equation}
which is to maximize the mutual information between $x$ and joint distribution of $z$ and $u$.

As Figure~\ref{fig:set} illustrates, we provide an intuitive demonstration of the optimal outcome of the feature extractor using a Venn diagram. Goal 1 is fulfilled as no overlap exist between $z$ (green area) and $u$ (orange area); and Goal 2 is achieved as $z$ and $u$ jointly (all colored regions) cover all the information of $x$ (blue circle).

It is widely accepted in the previous works that precisely calculating the mutual information between two arbitrary distributions are likely to be infeasible \cite{peng2018variational}. As a result, we replace the mutual information objectives in Equation~\ref{equ:min} and \ref{equ:altmax} with their upper and lower bounds for effective optimization. For Goal 1, we utilize the mutual information upper bound derived in~\cite{song2018learning} as:
\begin{equation}
\label{equ:upper}
    I(z;u) \leq \mathbb{E}_{q_\theta(z)}D_{KL}(q_\theta(u|z)||p(u)),
\end{equation}
for any distribution $p(u)$. 
Note that the term $q_\theta(u|z)$ in Equation~(\ref{equ:upper}) is hard to estimate and hence we instead replace the KL divergence term with its lower bound by introducing a conditional distribution $p_{\Psi}(u | z)$ parameterized with $\Psi$. It was shown in~\cite{song2018learning} that:
\begin{align}
    &\mathbb{E}_{q_\theta(z)}\left[\log p_{\Psi}(u | z) - \log p(u)\right] \nonumber\\ &\leq \mathbb{E}_{q_\theta(z)}D_{KL}(q_\theta(u|z)||p(u))
\end{align}
Hence, the Equation~\ref{equ:min} can be rewritten as an adversarial training objective function: 
\begin{equation}
    \label{equ:goal1}
    \min_\theta \max_\Psi \mathbb{E}_{q_\theta(z)}\left[\log p_{\Psi}(u | z) - \log p(u)\right],
\end{equation}
As $\log p(u)$ is a constant number that is independent of $z$, Equation~\ref{equ:goal1} can be further simplified to: 
\begin{equation}
    \label{equ:altgoal1}
    \max_\theta \min_\Psi -\mathbb{E}_{q_\theta(z)}\left[\log p_{\Psi}(u | z) \right],
\end{equation}
which is the cross entropy loss of predicting $u$ with $p_{\Psi}(u | z)$, i.e., $CE\left[p_{\Psi}(u | z)\right]$.
This objective function can be interpreted as an adversarial game between an adversary $p_{\Psi}$ who tries to infer $u$ from $z$ and a defender $q_{\theta}$ who aims to protect the user privacy.

For Goal 2, we adopt the previously proposed Jensen-Shannon mutual information estimator~\cite{nowozin2016f,hjelm2018learning} to estimate the lower bound of the mutual information I(x; z,u). The lower bound is formulated as follows:
\begin{align}
\label{eq:jsd}
    &\mathcal{I}(x;z,u)\ge \mathcal{I}_{\theta,\omega}^{(JSD)}(x;z,u) \nonumber\\
    &:=\mathbb{E}_{x}\left[-sp(-E_\omega(x;f_\theta(x),u))\right]-\mathbb{E}_{x,x'}\left[sp(E_\omega(x';f_\theta(x),u))\right],
\end{align}
where $x'$ is an random input data sampled independently from the same distribution of $x$, $sp(z)=log(1+e^z)$ is the softplus function and $E_\omega$ is a discriminator modeled by a neural network with parameters $\omega$. Hence, to maximally retain the original information, the feature extractor and the mutual information estimator can be optimized using Equation~\ref{eq:max_jsd}:
\begin{equation}
    \max_\theta \max_\omega \mathcal{I}_{\theta,\omega}^{(JSD)}(x;z,u)\label{eq:max_jsd}.
\end{equation} 
Finally, combining Equation~\ref{equ:altgoal1} and \ref{eq:max_jsd}, the objective function of training the feature extractor can be formulated as:
\begin{equation}
    \max_\theta(\lambda \min_\Psi CE\left[p_{\Psi}(u | z)\right] + (1-\lambda) \max_\omega\mathcal{I}_{\theta,\omega}^{(JSD)}(x;z,u)),
\end{equation}
where $\lambda \in [0,1]$ serves as a utility-privacy budget. A larger $\lambda$ indicates a stronger privacy protection, while a smaller $\lambda$ allowing more original information to be retained in the extracted features.
\section{Design of TIPRDC}\label{sec:framework_design}
\subsection{Overview}
The critical module of TIPRDC is the feature extractor. As presented in Section \ref{sec:problem_formulation}, there are two goals for learning the feature extractor, such that it can hide private attribute from features while retaining as much information of the raw data as possible to maintain the utility for primary learning task. To this end, we design a hybrid learning method to train the feature extractor, including the privacy adversarial training (PAT) algorithm and the MaxMI algorithm. In particular, we design the PAT algorithm, which simulates the game between an adversary who makes efforts to infer private attributes from the extracted features and a defender who aims to protect user privacy. By applying PAT to optimize the feature extractor, we enforce the feature extractor to hide private attribute $u$ from extracted features $z$, which is goal 1 introduced in Section \ref{sec:problem_formulation}. Additionally, we propose the MaxMI algorithm to achieve goal 2 presented in Section \ref{sec:problem_formulation}. By performing MaxMI to train the feature extractor, we can enable the feature extractor to maximize the mutual information between the information of the raw data $x$ and the joint information of the private attribute $u$ and the extracted feature $z$.

\begin{figure}[t]
\centering
    \captionsetup{width=1.0\linewidth}
		\includegraphics[scale=0.35]{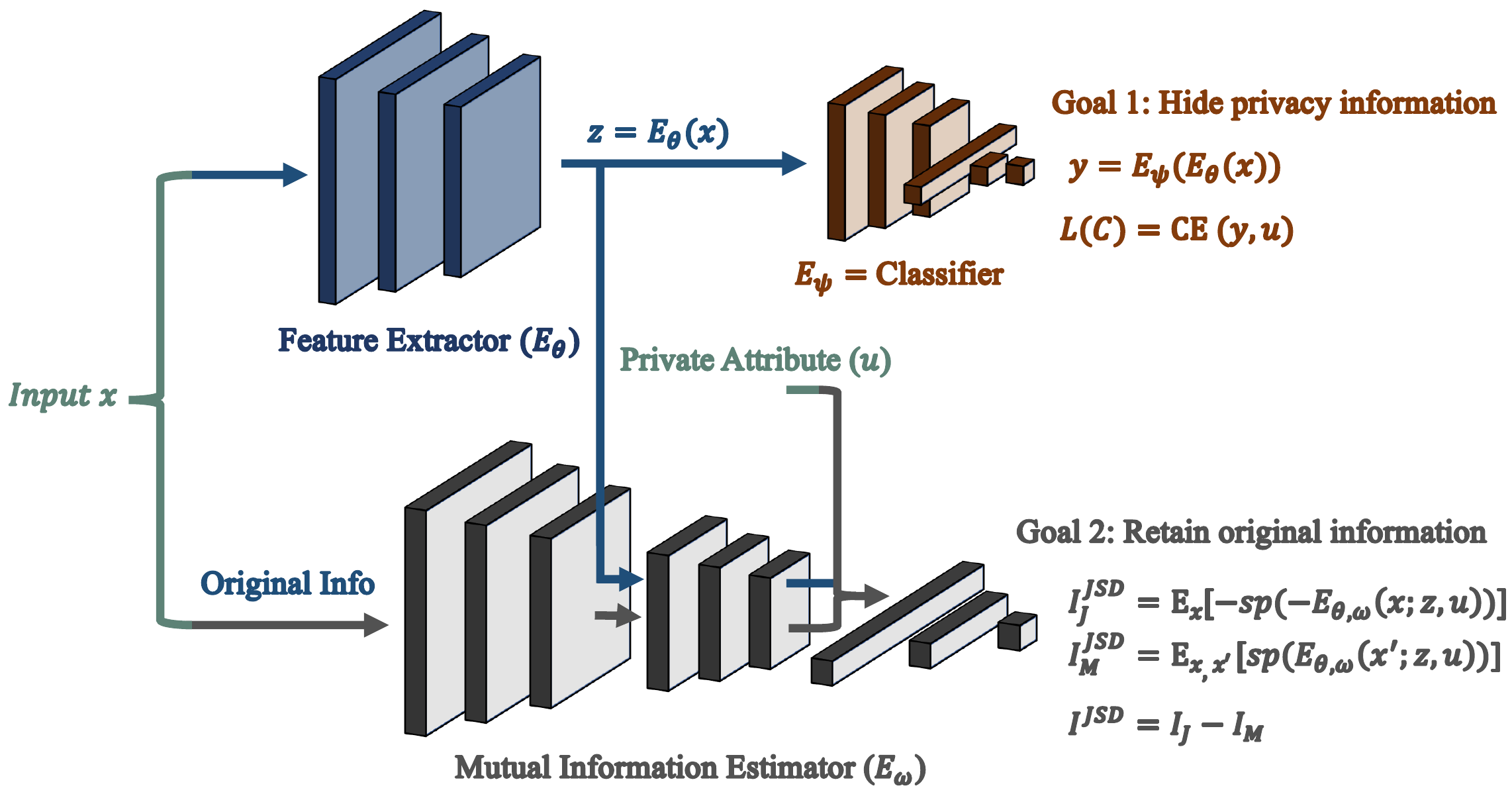}
	\caption{The hybrid learning method for training the feature extractor.}
	\label{fig:framework_design}
\end{figure}

As Figure \ref{fig:framework_design} shows, there are three neural network modules in the hybrid learning method: \textit{feature extractor}, \textit{adversarial classifier} and \textit{mutual information estimator}. The feature extractor is the one we aim to learn by performing the proposed hybrid learning algorithm. The adversarial classifier simulates an adversary in the PAT algorithm, aiming to infer private attribute $u$ from the eavesdropped features. The mutual information estimator is adopted in MaxMI algorithm to measure the mutual information between the raw data $x$ and the joint distribution of the private attribute $u$ and the extracted feature $z$. All three modules are end-to-end trained using our proposed hybrid learning method.

Before presenting the details of each algorithm, we give the following notations. As same as presented in Section \ref{sec:problem_formulation}, we adopt $x$, $u$ and $z$ as the raw data, the private attribute and the extracted feature, respectively. We denote $E_\theta$ as the feature extractor that is parameterized with $\theta$, and then $z$ can be expressed as $E_\theta(x)$. Let $E_\Psi$ represent the classifier, where $\Psi$ indicates the parameter set of the classifier. We adopt $y=E_\Psi(E_\theta(x))$ to denote the prediction generated by $E_\Psi$. Let $E_\omega$ denotes the mutual information estimator, where $\omega$ represents the parameter set of the mutual information estimator. 

\subsection{Privacy Adversarial Training Algorithm}
We design the PAT algorithm to achieve goal 1, enabling the feature extractor to hide $u$ from $E_\theta(x)$. The PAT algorithm is designed by simulating the game between an adversary who makes efforts to infer private attributes from the extracted features and a defender who aims to protect user privacy. We can apply any architecture to both the feature extractor and the classifier based on the requirement of data format and the primary learning task. The performance of the classifier ($C$) is measured using the cross-entropy loss function as:
\begin{equation}
    \mathcal{L}(C)=CE(y,u)=CE(E_\Psi(E_\theta(x),u),
\end{equation}
where $CE\left[\cdot\right]$ stands for the cross entropy loss function. When we simulate an adversary who tries to enhance the accuracy of the adversary classifier as high as possible, the classifier needs to be optimized by minimizing the above loss function as:
\begin{equation}
\label{eq:adv_lc}
    \Psi=\underset{\Psi}{\arg\min} \mathcal{L}(C).
\end{equation}
On the contrary, when defending against private attribute leakage, we train the feature extractor in PAT that aims to degrade the performance of the classifier. Consequently, the feature extractor can be trained using Equation \ref{eq:lc} when simulating a defender: 
\begin{equation}
\label{eq:lc}
    \theta=\underset{\theta}{\arg\max} \mathcal{L}(C).
\end{equation}
Based on Equation \ref{eq:adv_lc} and \ref{eq:lc}, the feature extractor and the classifier can be jointly optimized using Equation \ref{eq:pat}:
\begin{equation}
\label{eq:pat}
    \theta,\Psi=\arg\max_\theta \min_\Psi \mathcal{L}(C) ,
\end{equation}
which is consistent with Equation \ref{equ:altgoal1}.

\subsection{MaxMI Algorithm}
For goal 2, we propose MaxMI algorithm to make the feature extractor retain as much as information from the raw data as possible, in order to maintain the high utility of the extracted features. Specifically, the Jensen-Shannon mutual information estimator~\cite{nowozin2016f,hjelm2018learning} is adopted to measure the lower bound of the mutual information $I(x; z,u)$. Here we adopt $E_\omega$ as the the Jensen-Shannon mutual information estimator, and we can rewrite Equation \ref{eq:jsd} as:
\begin{align}
    &\mathcal{I}(x;z,u)\ge \mathcal{I}_{\theta,\omega}^{(JSD)}(x;z,u) \nonumber\\
    &:=\mathbb{E}_{x}\left[-sp(-E_\omega(x,E_\theta(x),u))\right]-\mathbb{E}_{x,x'}\left[sp(E_\omega(x',E_\theta(x),u))\right],
\end{align}
where $x'$ is an random input data sampled independently from the same distribution of $x$, $sp(z)=log(1+e^z)$ is the softplus function. Hence, to maximally retain the original information, the feature extractor and the mutual information estimator can be optimized using Equation~\ref{eq:new_jsd}:
\begin{align}
    \theta,\omega&=\arg\max_\theta \max_\omega \mathcal{I}_{\theta,\omega}^{(JSD)}(x;z,u), \label{eq:new_jsd}
\end{align} 
which is consistent with Equation \ref{eq:max_jsd}. Considering the difference of $x$, $z$ and $u$ in dimensionality, we feed them into the $E_\omega$ from different layers as illustrated in Figure \ref{fig:framework_design}. For example, the private attribute $u$ may be a binary label, which is represented by one bit. However, $x$ may be high dimensional
data (e.g., image), and hence it is not reasonable feed both $x$ and $u$ from the first layer of $E_\omega$.

\subsection{Hybrid Learning Method}
Finally, the feature extractor is trained by alternatively performing PAT algorithm and MaxMI algorithm. As aforementioned, we also introduce an utility-privacy budget $\lambda$ to balance the tradeoff between protecting privacy and retaining the original information. Therefore, combining Equation~\ref{eq:pat} and \ref{eq:new_jsd}, the objective function of training the feature extractor can be formulated as:
\begin{equation}
     \theta,\Psi,\omega=\arg\max_\theta(\lambda \min_\Psi \mathcal{L}(C) + (1-\lambda) \max_\omega\mathcal{I}_{\theta,\omega}^{(JSD)}(x;z,u)). \label{eq:hybrid}
\end{equation}

Algorithm~\ref{Algo:TIPRDC} summarizes the hybrid learning method of TIPRDC. Before performing the hybrid learning, we first jointly pretrain the feature extractor and the adversarial classifier normally without adversarial objective to obtain the best performance on classifying a specific private attribute. Then within each training batch, we first perform PAT algorithm and MaxMI algorithm to update $\Psi$ and $\omega$, respectively. Then, the feature extractor will be updated according to Equation \ref{eq:hybrid}.

\begin{algorithm}
    \caption{Hybrid Learning Method}
    \label{Algo:TIPRDC}
    \begin{algorithmic}[1]
        \Require
        Dataset $\mathcal{D}$
        \Ensure
        $\theta$
        \For{every epoch}
            \For{every batch}
                \State $\mathcal{L}(C)\rightarrow$ update $\psi$ (performing PAT)
                \State $-\mathcal{I}_{\theta, \omega}^{JSD}(x;z,u)\rightarrow$ update $\omega$ (performing MaxMI)
                \State $-\lambda\mathcal{L}(C)-(1-\lambda)\mathcal{I}_{\theta, \omega}^{JSD}(x;z,u)\rightarrow$ update $\theta$
            \EndFor
        \EndFor
    \end{algorithmic}
\end{algorithm}

\section{Evaluation}\label{sec:evaluation}
In this section, we evaluate TIPRDC's performance on three real-world datasets, with a focus on the utility-privacy tradeoff. We also compare TIPRDC with existing solutions proposed in the literature and visualize the results. 

\subsection{Experiment Setup}
We evaluate TIPRDC, especially the learned feature extractor, on two image datasets and one text dataset. We implement TIPRDC with PyTorch, and train it on a server with 4$\times$NVIDIA TITAN RTX GPUs. We apply mini-batch technique in training with a batch size of 64, and adopt the AdamOptimizer \cite{adam} with an adaptive learning rate in the hybrid learning procedure. The architecture configurations of each module are presented in Table \ref{tb:model_arch_img} and \ref{tb:model_arch_text}. For evaluating the performance, given a primary learning task, a simulated data collector trains a normal classifier using features processed by the learned feature extractor, and such normal classifier has the same architecture of the classifier presented in Table \ref{tb:model_arch_img} and \ref{tb:model_arch_text}. The utility and privacy of the extracted features $E_\theta(x)$ are evaluated by the classification accuracy of primary learning tasks and specified private attribute, respectively.

We adopt CelebA \cite{liu2015faceattributes}, LFW \cite{kumar2009attribute} and the dialectal tweets dataset (DIAL) \cite{blodgett2016demographic} for the training and testing of TIPRDC. CelebA consists of more than 200K face images. Each face image is labeled with 40 binary facial attributes. The dataset is split into 160K images for training and 40K images for testing. LFW consists of more than 13K face images, and each face image is labeled with 16 binary facial attributes. We split LFW into 10K images for training and 3K images for testing. DIAL consists of 59.2 million tweets collected from 2.8 million users, and each tweet is annotated with three binary attributes. DIAL is split into 48 million tweets for training and 11.2 million tweets for testing.

\begin{table}[h]
\caption{The architecture configurations of each module for CelebA and LFW.}\vspace{-0.15in}
    \begin{tabular}{l|l|l}
    \hline
    \textbf{Feature Extractor} & \textbf{Classifier} & \textbf{MI Estimator}                                           \\ \hline
    2$\times$conv3-64          & 3$\times$conv3-256  & 3$\times$conv3-64                                               \\
    maxpool                    & maxpool             & maxpool                                                         \\ \hline
    2$\times$conv3-128         & 3$\times$conv3-512  & 2$\times$conv3-128                                              \\
    maxpool                    & maxpool             & maxpool                                                         \\ \hline
    \textbf{}                  & 3$\times$conv3-512  & 3$\times$conv3-256                                              \\
                               & maxpool             & maxpool                                                         \\ \cline{2-3} 
                               & 2$\times$fc-4096    & 3$\times$conv3-512                                              \\
                               & fc-label length     & maxpool                                                         \\ \cline{2-3} 
                               &                     & 3$\times$conv3-512                                              \\
                               &                     & maxpool                                                         \\ \cline{3-3} 
                               &                     & \begin{tabular}[c]{@{}l@{}}fc-4096\\ fc-512\\ fc-1\end{tabular} \\ \hline
    \end{tabular}\label{tb:model_arch_img}
\end{table}\vspace{-0.2in}

\begin{table}[h]
\caption{The architecture configurations of each module for DIAL.}\vspace{-0.15in}
    \begin{tabular}{l|l|l}
    \hline
    \textbf{Feature Extractor} & \textbf{Classifier} & \textbf{MI Estimator} \\ \hline
    embedding-300              & 2$\times$lstm-300    & embedding-300         \\ \hline
    lstm-300                   & fc-150              & lstm-300              \\ \cline{1-1} \cline{3-3} 
                               & fc-label length     & 2$\times$lstm-300      \\ \cline{2-3} 
                               &                     & fc-150                \\
    \textbf{}                  &                     & fc-1                  \\ \hline
    \end{tabular}\label{tb:model_arch_text}
\end{table}\vspace{-0.2in}

\subsection{Comparison Baselines}
We select four types of data privacy-preserving baselines \cite{liu2019privacy}, which have been widely applied in the literature, and compare them with TIPRDC. The details settings of the baseline solutions are presented as below.
\begin{itemize}
    \item \textbf{Noisy} method perturbs the raw data $x$ by adding Gaussian noise $\mathcal{N}(0,\sigma^2)$, where $\sigma$ is set to 40 according to \cite{liu2019privacy}. The noisy data $\Bar{x}$ will be delivered to the data collector. The Gaussian noise injected to the raw data can provide strong guarantees of differential privacy using less local noise. This scheme has been widely applied in federated learning \cite{papernot2018scalable,truex2019hybrid}. 
    \item \textbf{DP} approach injects Laplace noise the raw data $x$ with diverse privacy budgets \{0.1, 0.2, 0.5, 0.9\}, which is a typical differential privacy method. The noisy data $\Bar{x}$ will be submitted to the data collector. 
    \item \textbf{Encoder} learns the latent representation of the raw data $x$ using a DNN-based encoder. The extracted features $z$ will be uploaded to the data collector.
    \item \textbf{Hybrid} method \cite{osia2020hybrid} further perturbs the above encoded features by performing principle components analysis (PCA) and adding Laplace noise  with varying noise factors privacy budgets \{0.1, 0.2, 0.5, 0.9\}.
\end{itemize}

\subsection{Evaluations on CelebA and LFW}
\begin{figure*}[t]
    \centering
    \subfigure[]{\includegraphics[scale=0.26]{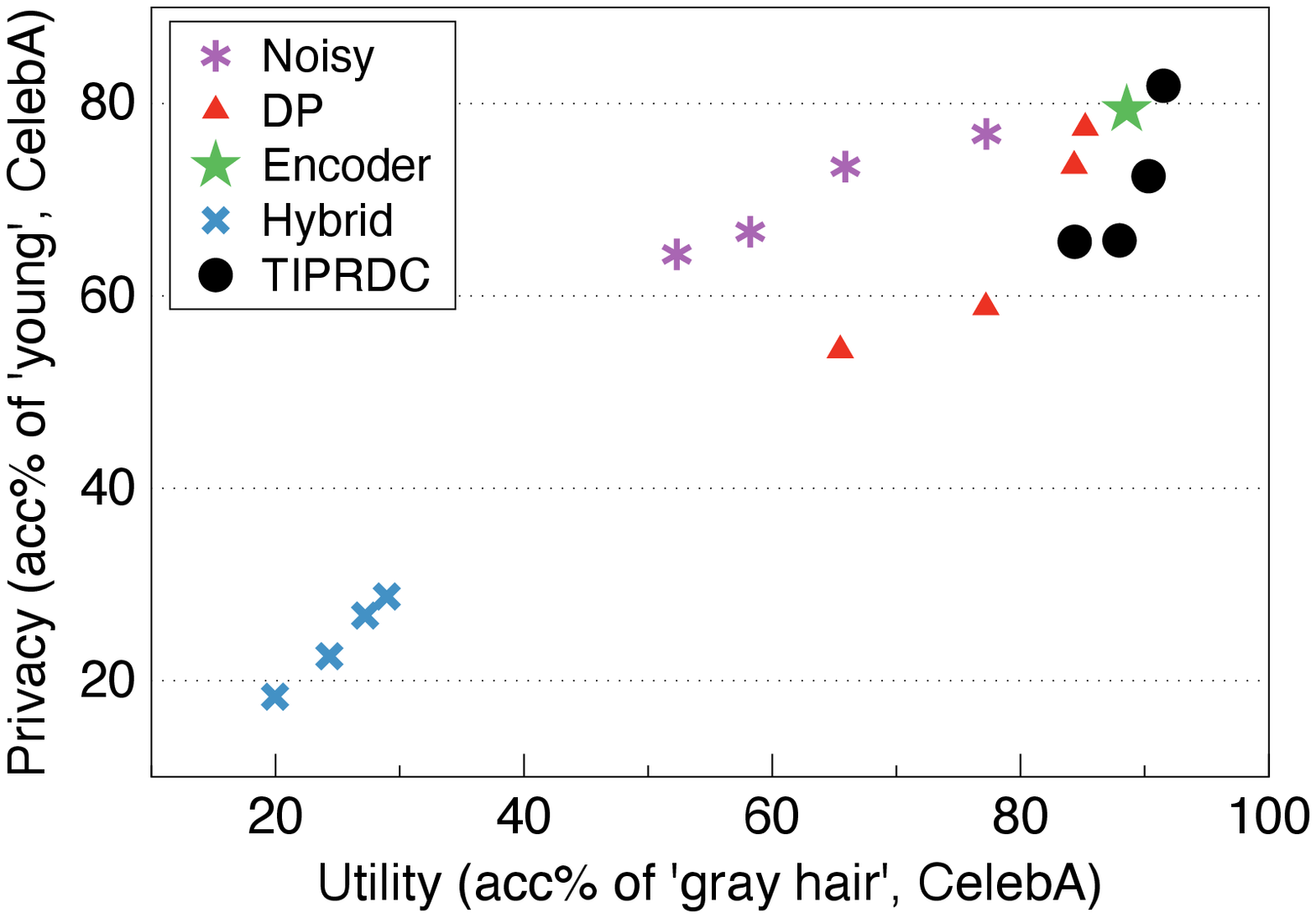}}
    \subfigure[]{\includegraphics[scale=0.26]{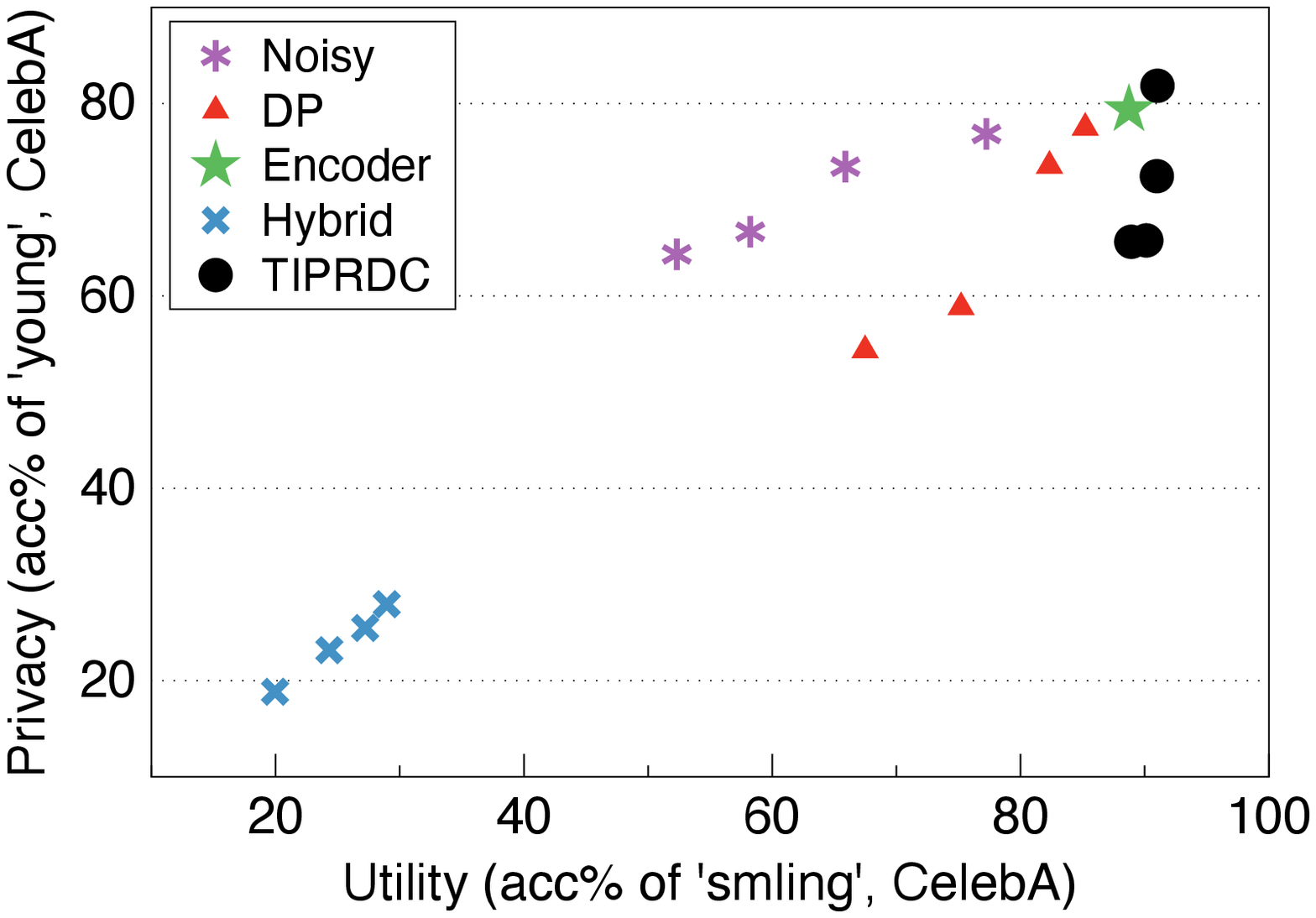}}
    \subfigure[]{\includegraphics[scale=0.26]{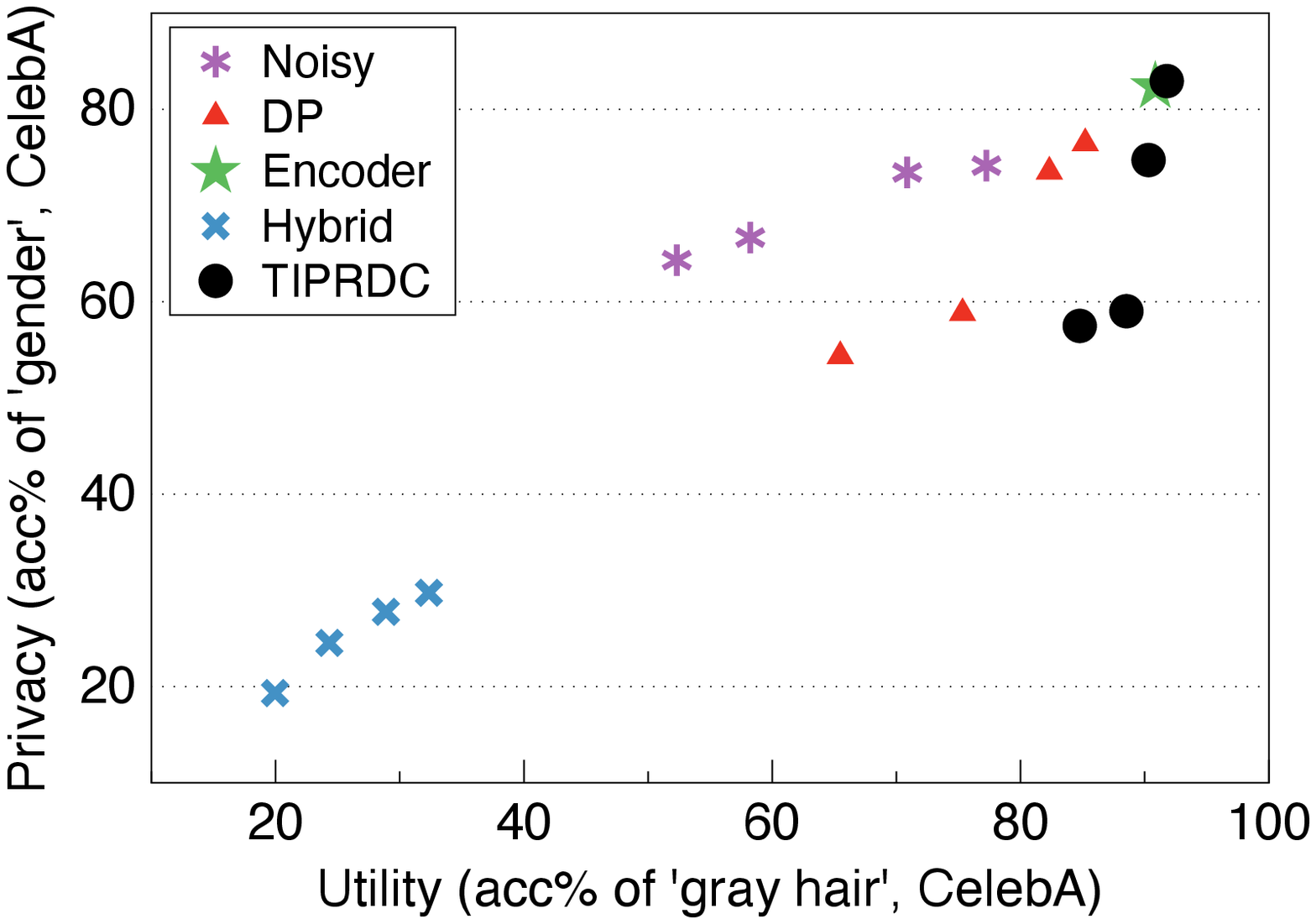}}
    \subfigure[]{\includegraphics[scale=0.26]{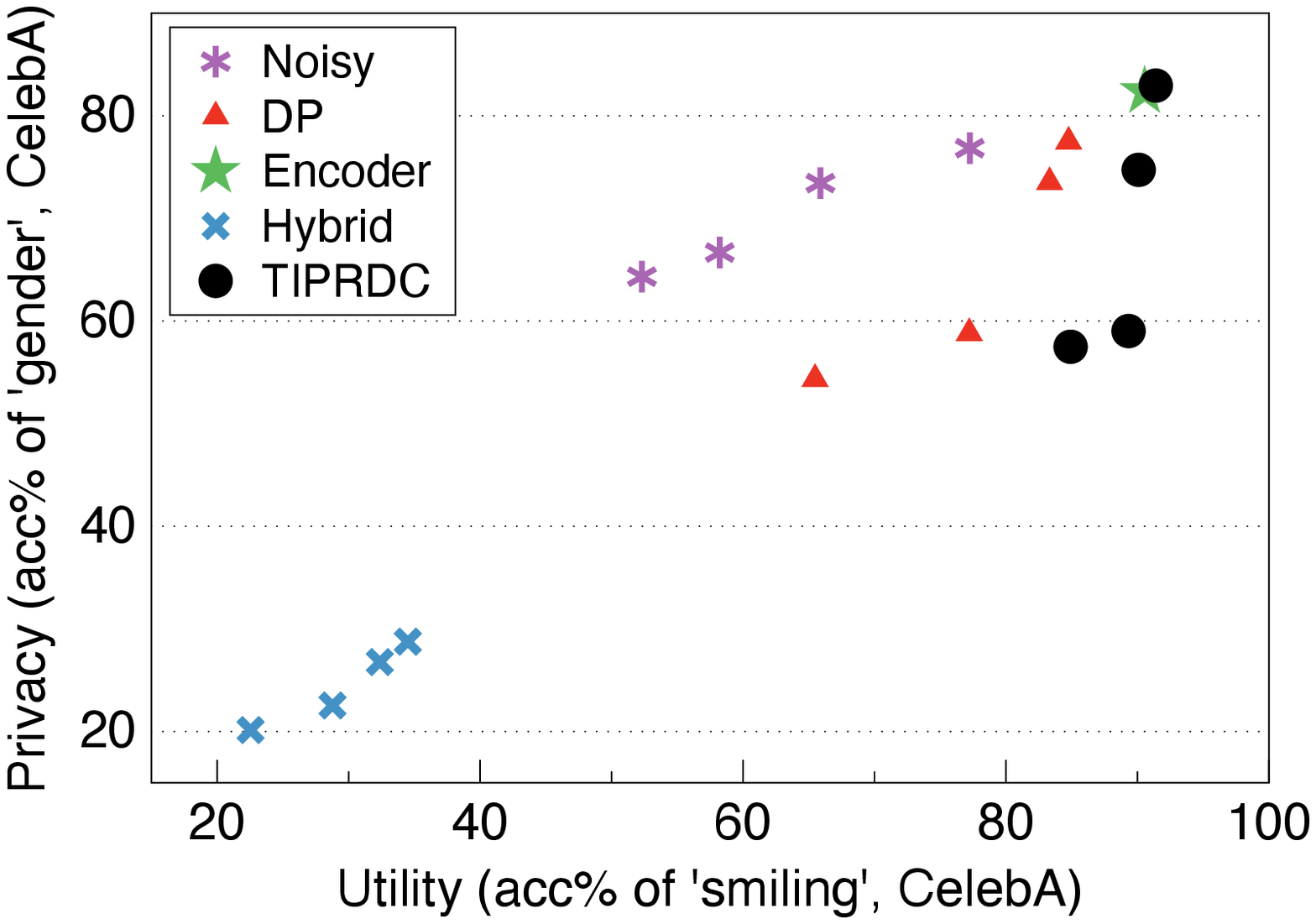}}
    \subfigure[]{\includegraphics[scale=0.26]{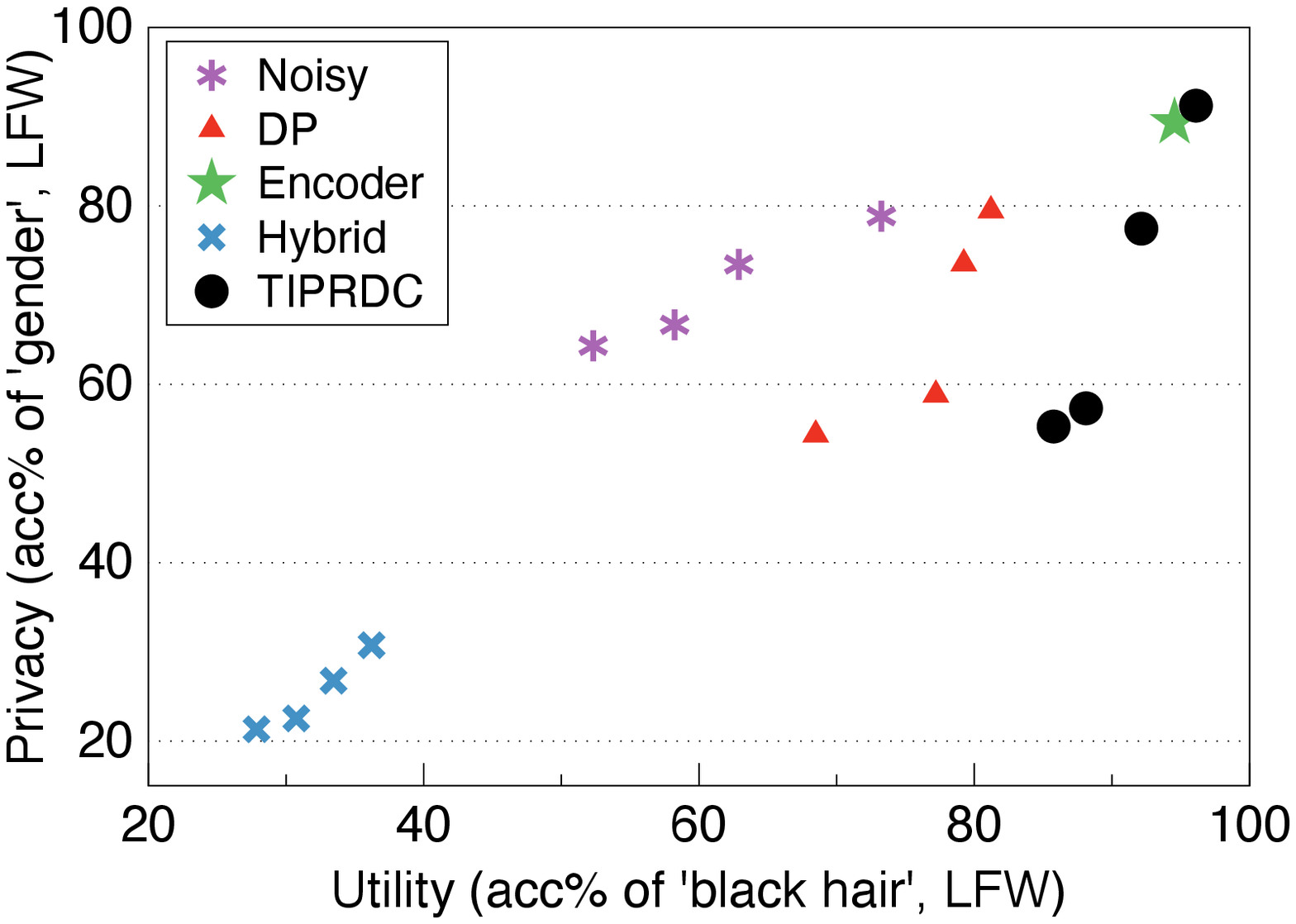}}
    \subfigure[]{\includegraphics[scale=0.26]{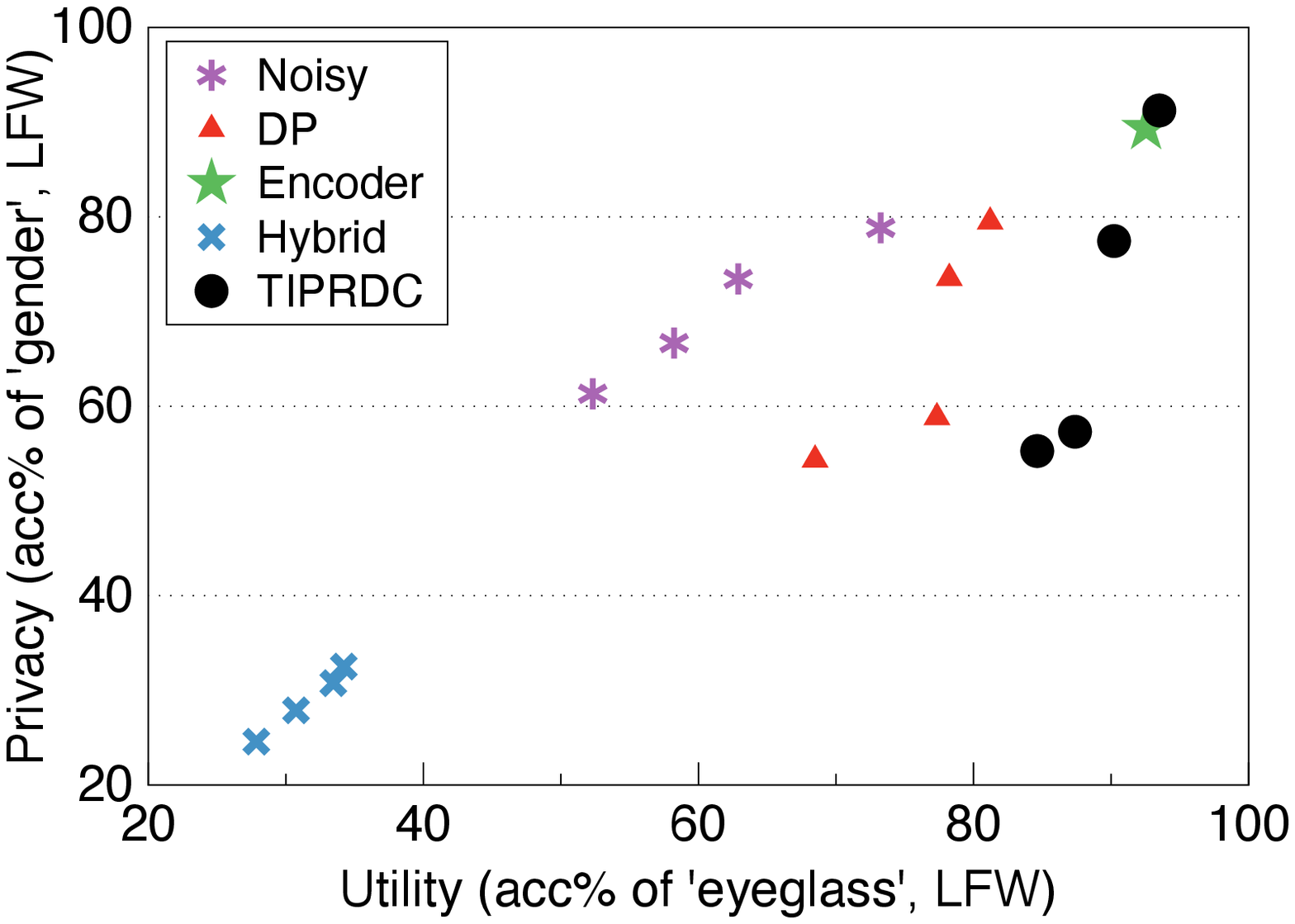}}
    \subfigure[]{\includegraphics[scale=0.26]{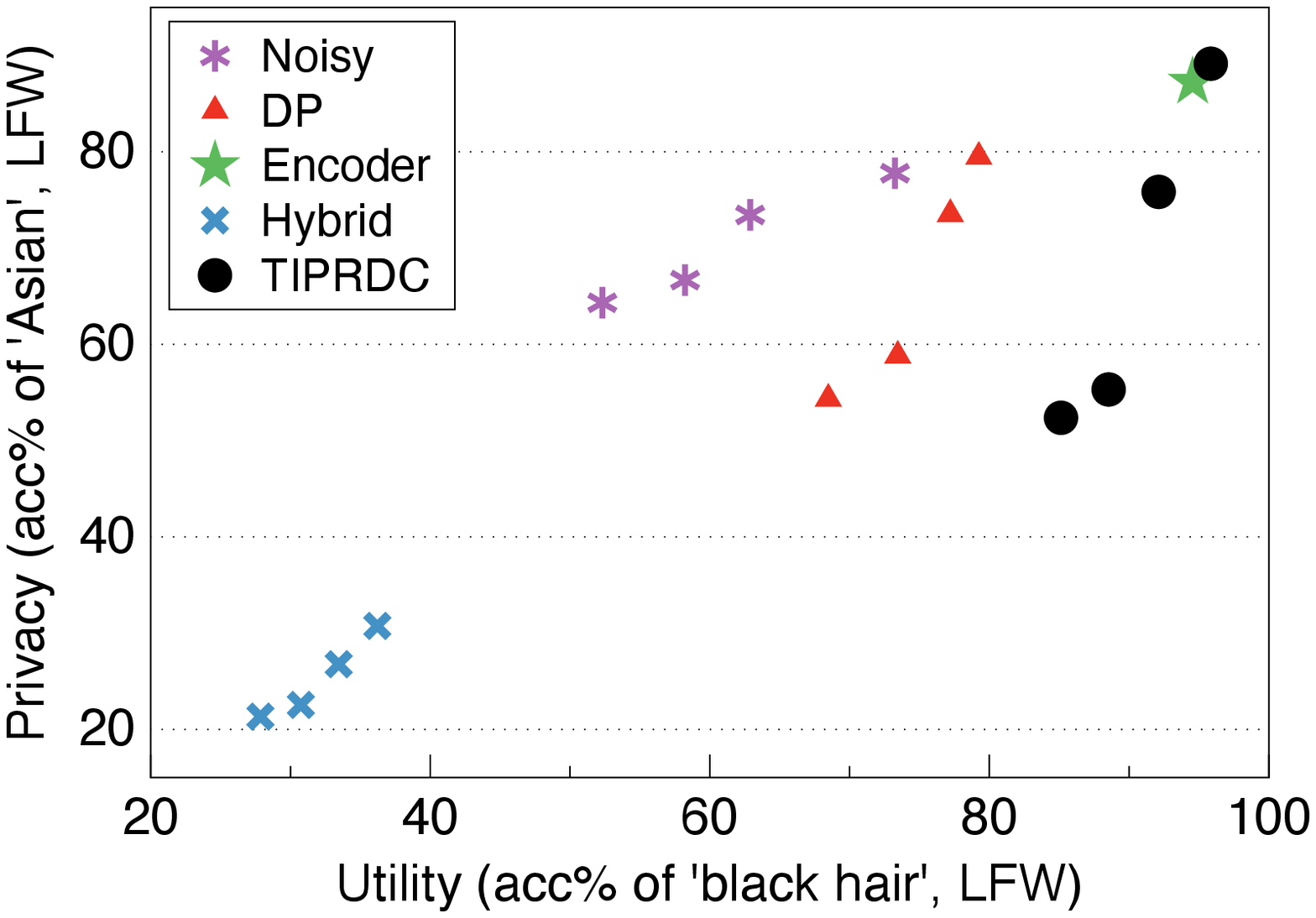}}
    \subfigure[]{\includegraphics[scale=0.26]{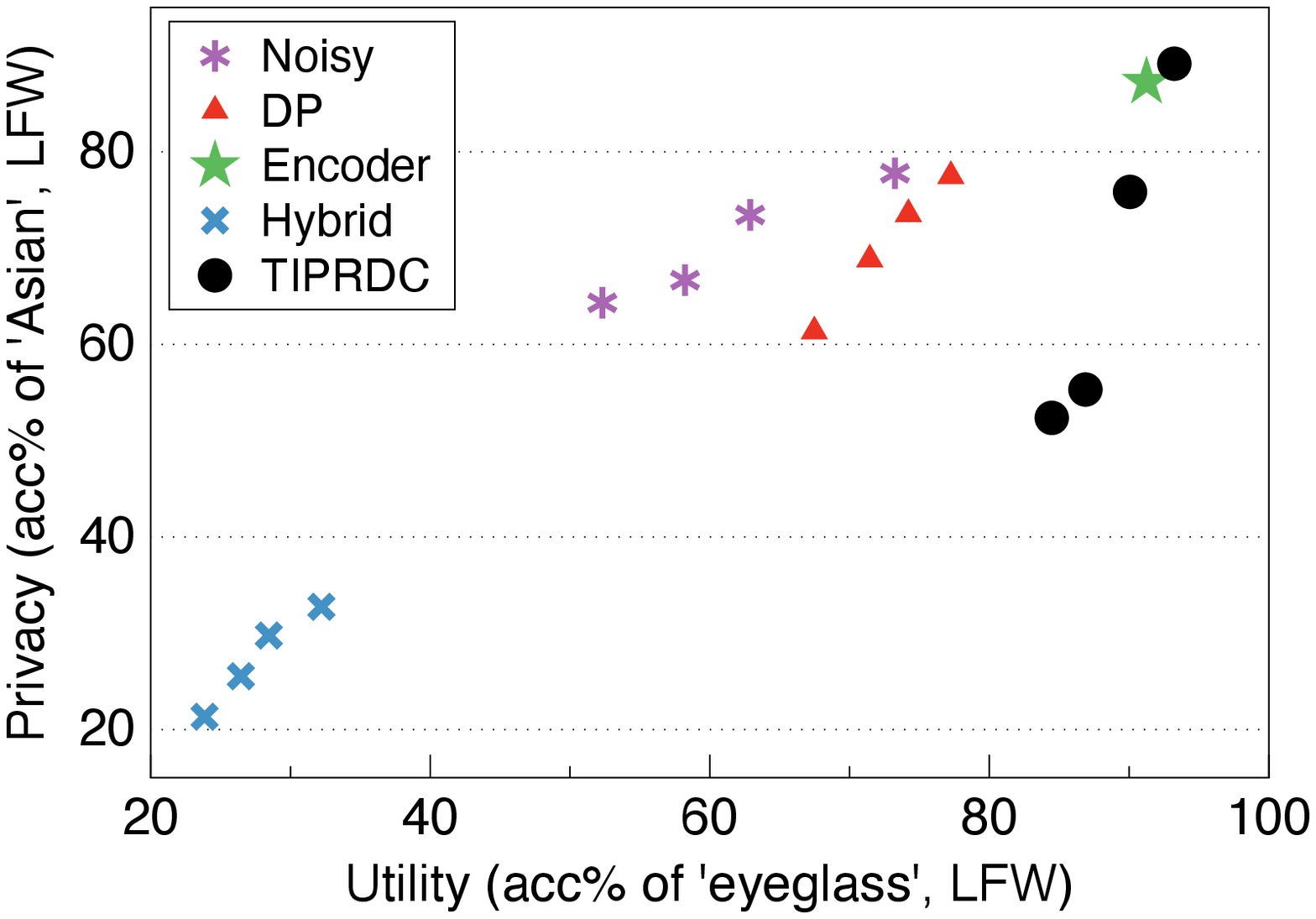}}\vspace{-0.2in}
    \caption{Utility-privacy tradeoff comparison of TIPRDC with four baselines on CelebA and LFW.}
    \label{fig:comparsion}
\end{figure*}

\textbf{Comparison of utility-privacy tradeoff:} We compare the utility-privacy tradeoff offered by TIPRDC with four privacy-preserving baselines. 
In our experiments, we set `young' and `gender' as the private labels to protect in CelebA, and consider detecting `gray hair' and `smiling' as the primary learning tasks to evaluate the utility. 
With regard to LFW, we set `gender' and `Asian' as the private labels, and choose recognizing `black hair' and `eyeglass' as the classification tasks. 
Figure \ref{fig:comparsion} summarizes the utility-privacy tradeoff offered by four baselines and TIPRDC. Here we evaluate TIPRDC with four discrete choices of $\lambda\in\{1, 0.9, 0.5, 0\}$.

As Figure \ref{fig:comparsion} shows, although TIPRDC cannot always outperform the baselines in both utility and privacy, it still achieve the best utility-privacy tradeoff under most experiment settings. 
For example, in Figure \ref{fig:comparsion} (h), TIPRDC achieves the best tradeoff by setting $\lambda=0.9$. Specifically, the classification accuracy of `Asian' on LFW is 55.31\%, and the accuracy of `eyeglass' is 86.88\%. This demonstrates that TIPRDC can efficiently protect privacy while maintaining high utility of extracted features. 

In other four baselines, Encoder method can maintain a good utility of the extracted features, but it fails to protect privacy due to the high accuracy of private labels achieved by the adversary. Noisy, DP and Hybrid methods offer strong privacy protection with sacrificing the utility.

\textbf{Impact of the utility-privacy budget $\lambda$:} An important step in the hybrid learning procedure (see Equation \ref{eq:hybrid}) is to determine the utility-privacy budget $\lambda$. 
To determine the optimal $\lambda$, we evaluate the utility-privacy tradeoff on CelebA and LFW by setting different $\lambda$. Specifically, we evaluate the impact of $\lambda$ with four discrete choices of $\lambda\in\{1, 0.9, 0.5, 0\}$. The private labels and primary learning tasks in CelebA and LFW are set as same as the above experiments.

\begin{figure*}[t]
    \centering
    \subfigure[CelebA]{\includegraphics[scale=0.215]{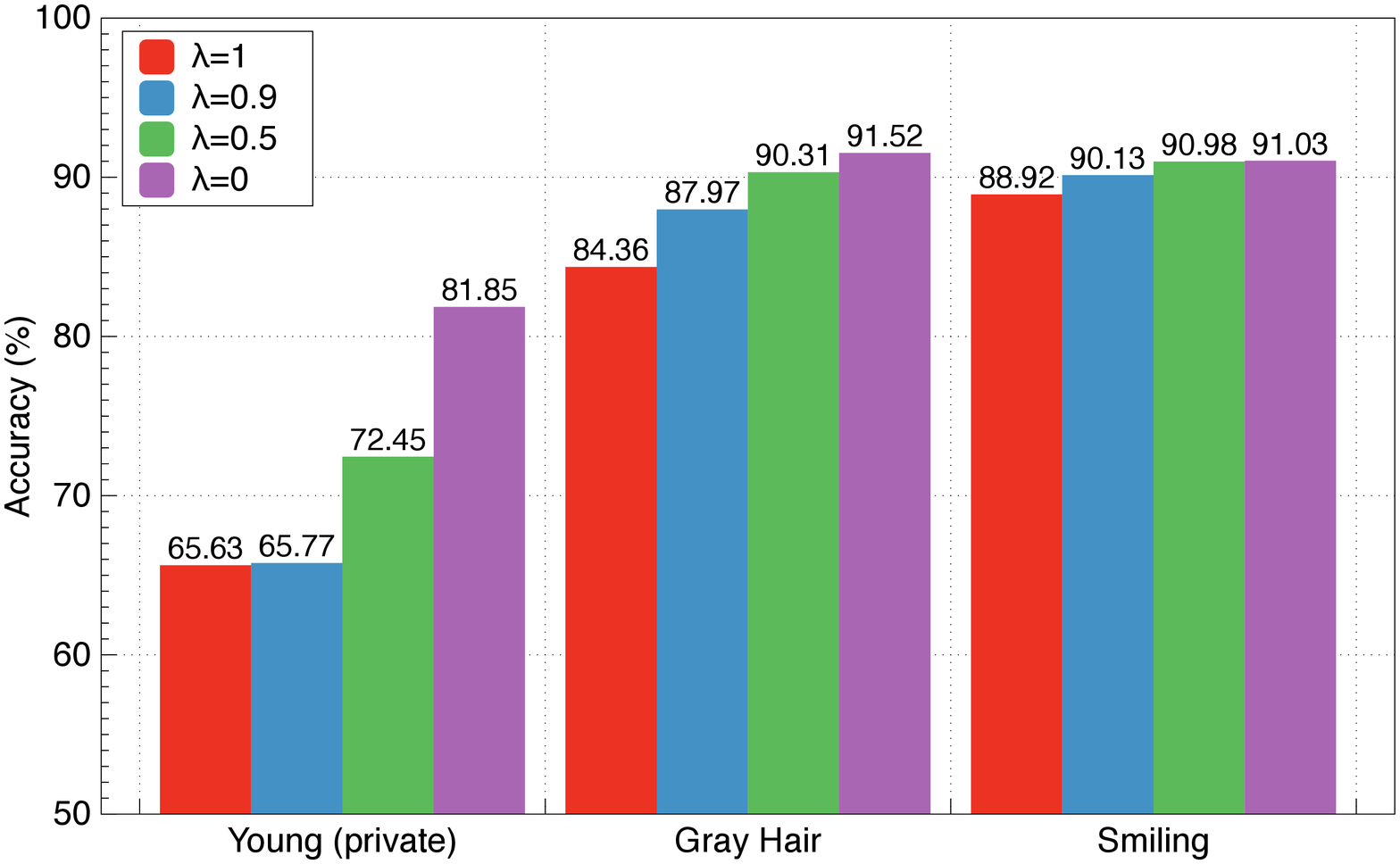}}
    \subfigure[CelebA]{\includegraphics[scale=0.215]{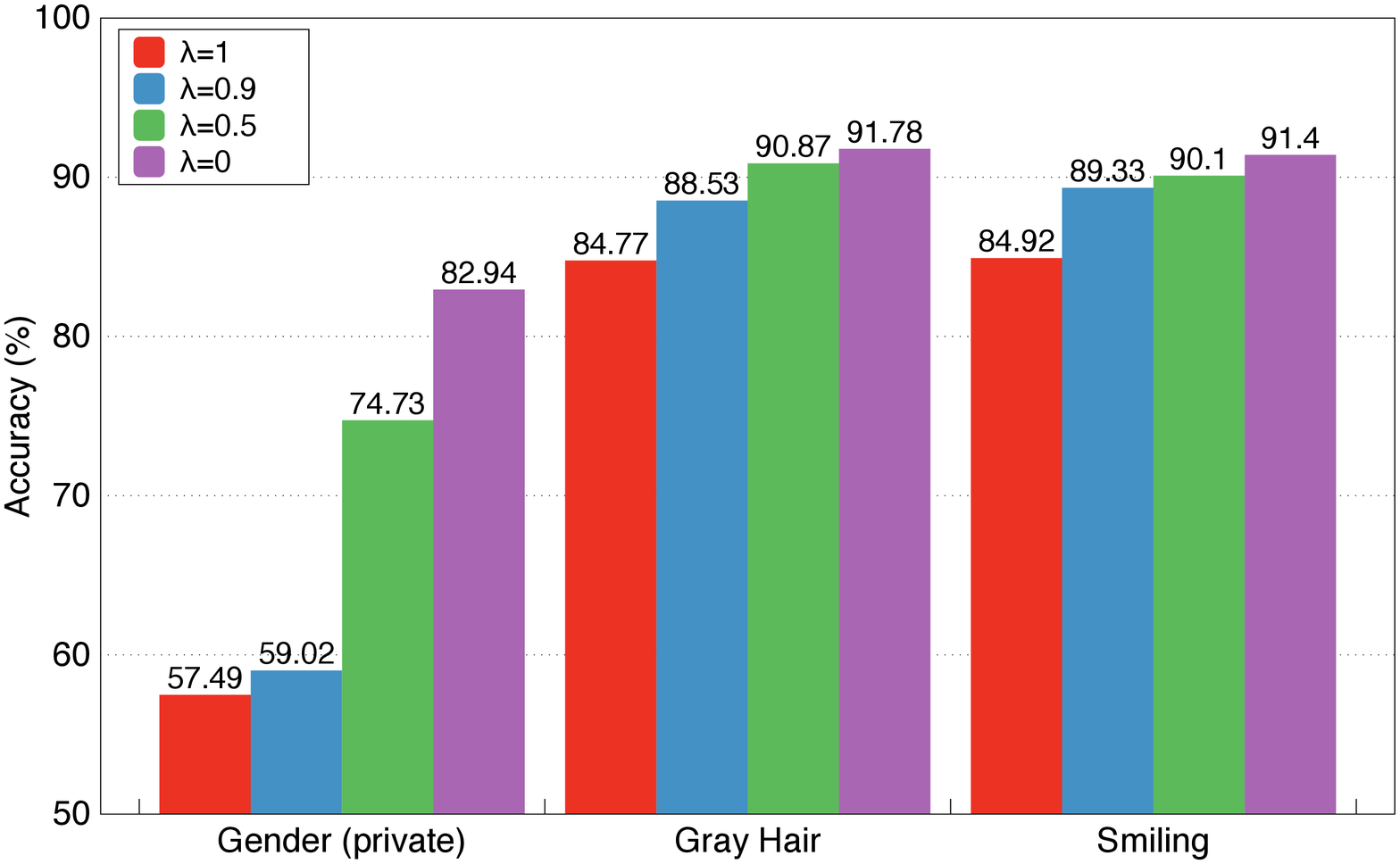}}
    \subfigure[CelebA]{\includegraphics[scale=0.215]{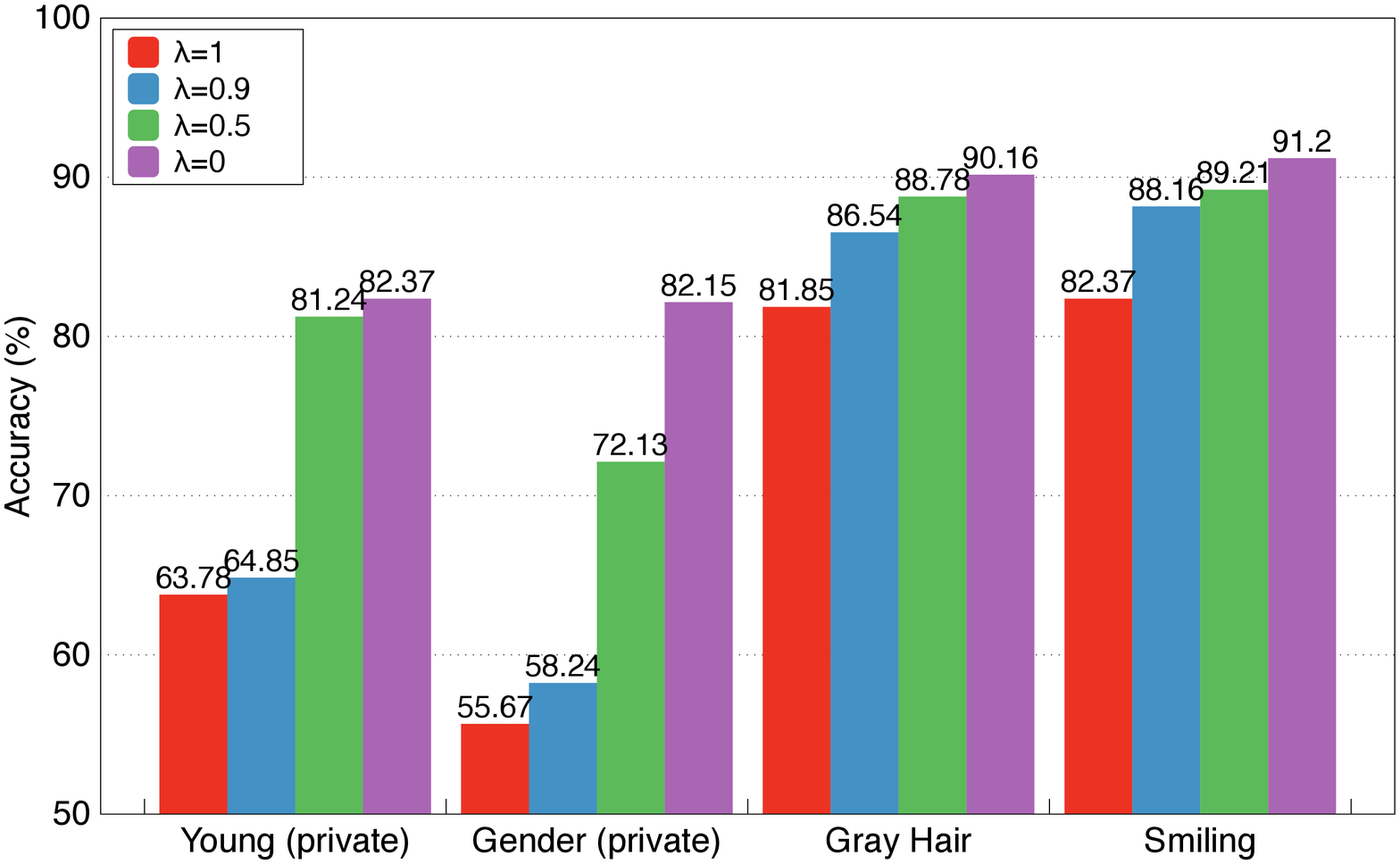}}
    \subfigure[LFW]{\includegraphics[scale=0.215]{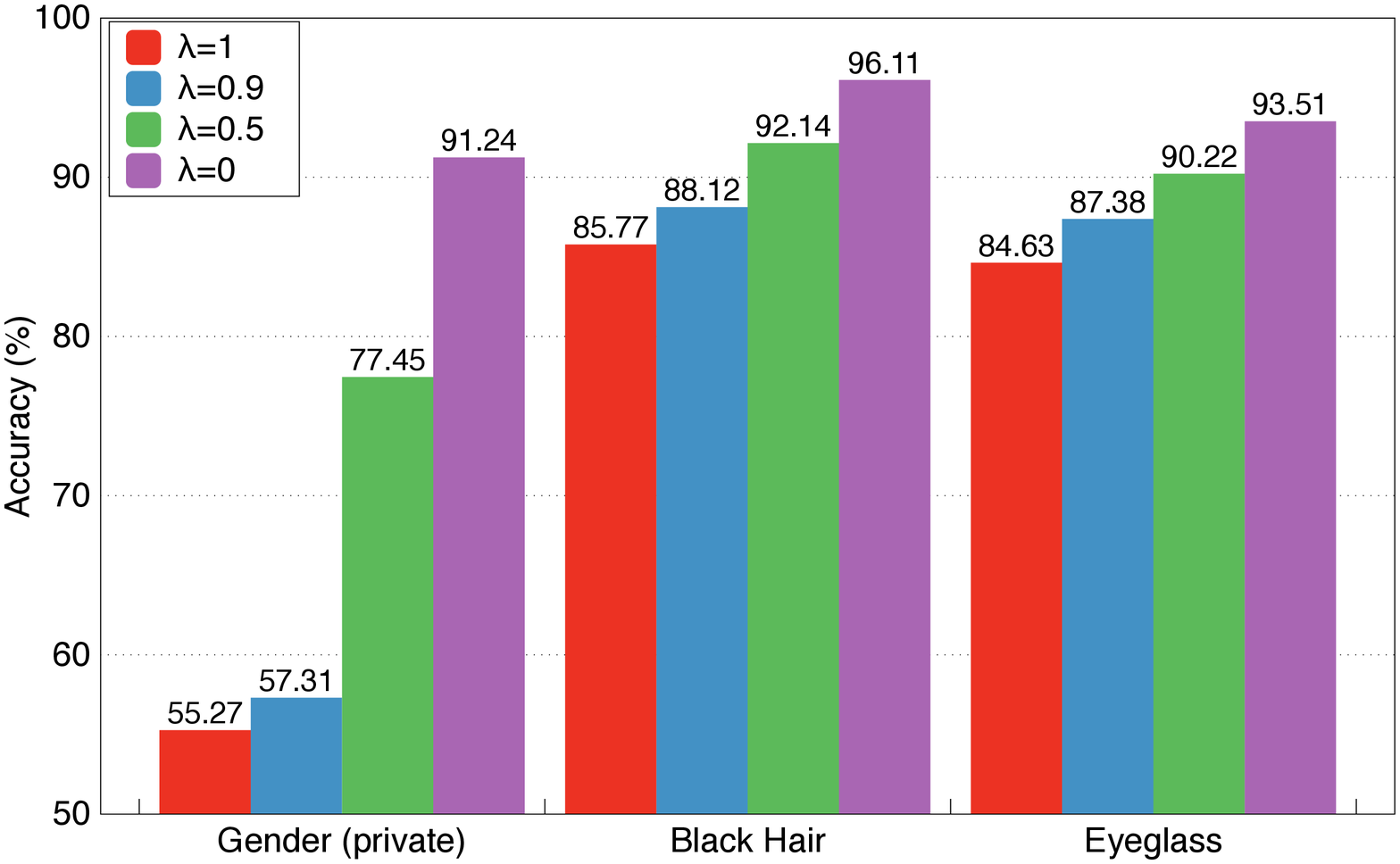}}
    \subfigure[LFW]{\includegraphics[scale=0.215]{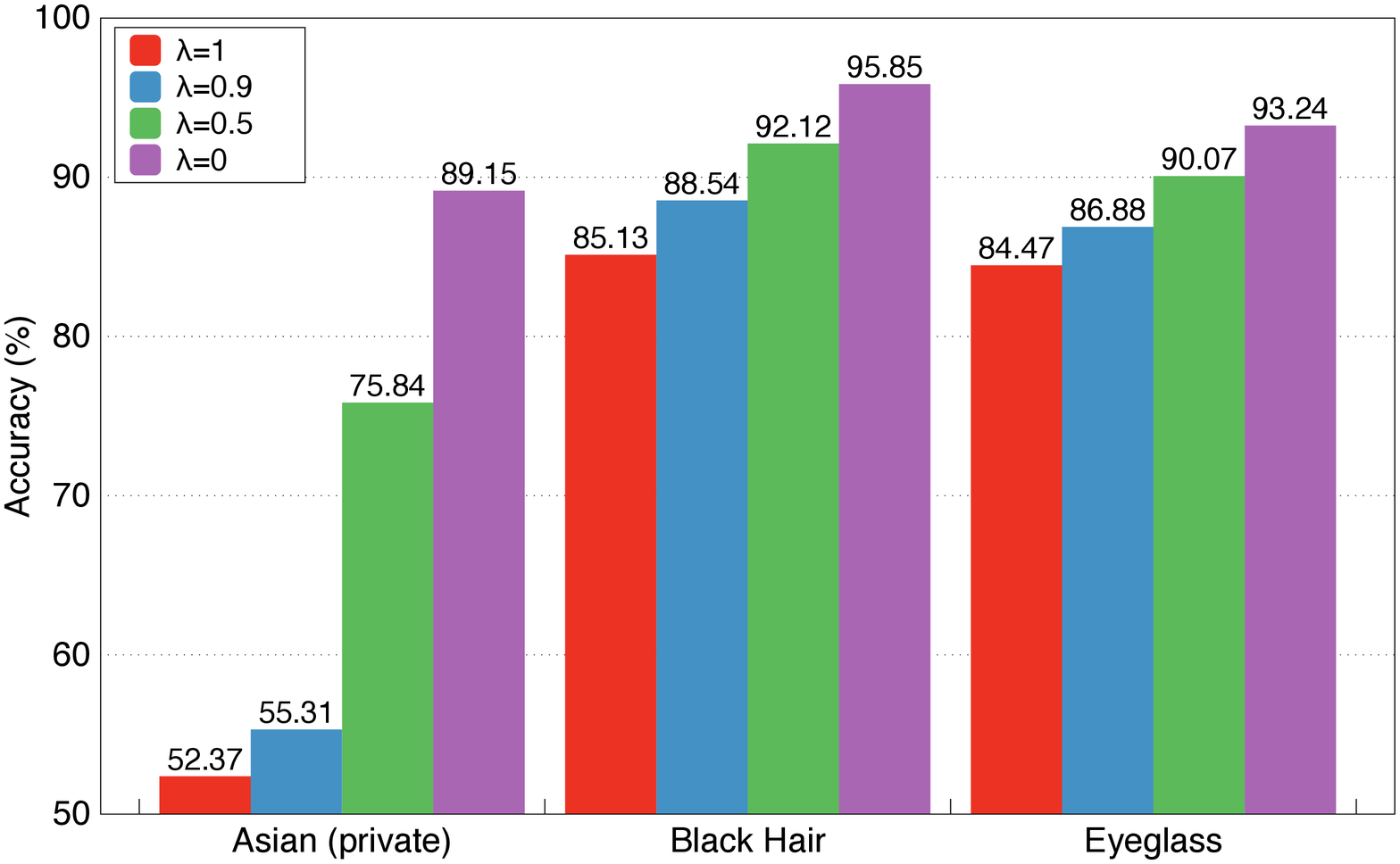}}
    \subfigure[LFW]{\includegraphics[scale=0.215]{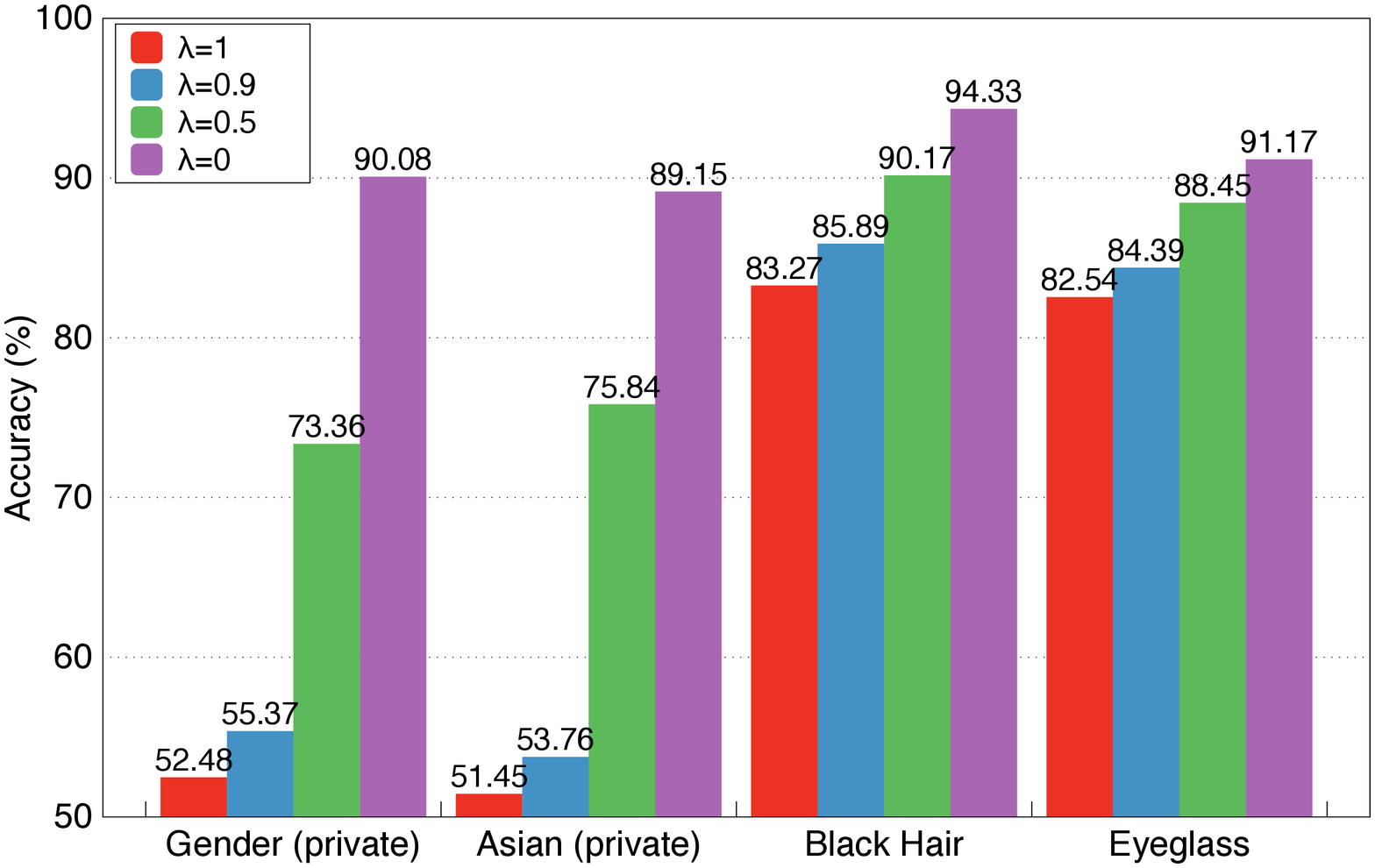}}\vspace{-0.2in}
    \caption{The impact of the utility-privacy budget $\lambda$ on CelebA and LFW.}
    \label{fig:budget}
\end{figure*}

As Figure \ref{fig:budget} illustrates, the classification accuracy of primary learning tasks will increase with a smaller $\lambda$, but the privacy protection will be weakened. Such phenomenon is reasonable, since the smaller $\lambda$ means hiding less privacy information in features but retaining more original information from the raw data according to Equation \ref{eq:hybrid}. For example, in Figure \ref{fig:budget} (a), the classification accuracy of `gray hair' on CelebA is 84.36\% with $\lambda=1$ and increases to 91.52\% by setting $\lambda=0$; the classification accuracy of `young' is 65.63\% and 81.85\% with decreasing $\lambda$ from 1 to 0, respectively. Overall, $\lambda=0.9$ is an optimal utility-privacy budget for experiment settings in both CelebA and LFW.

\begin{figure}[t]
    \centering
    \subfigure[raw image]{\includegraphics[scale=0.3]{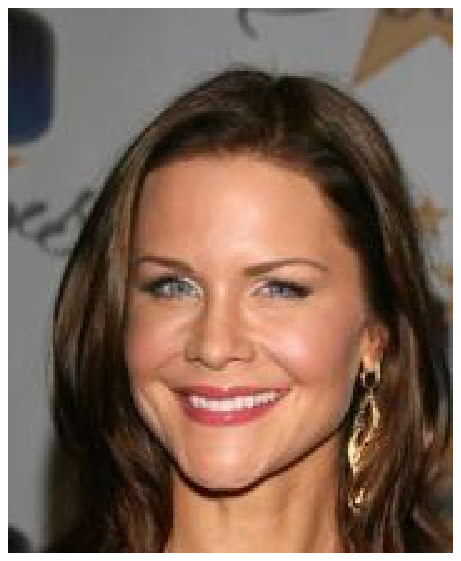}}
    \subfigure[$\lambda=1$]{\includegraphics[scale=0.3]{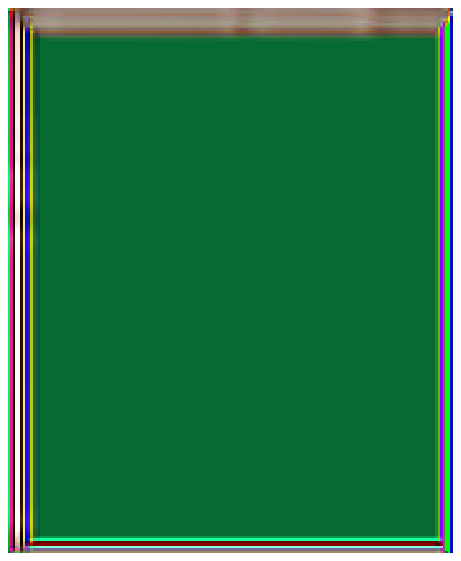}}
    \subfigure[$\lambda=0.9$]{\includegraphics[scale=0.3]{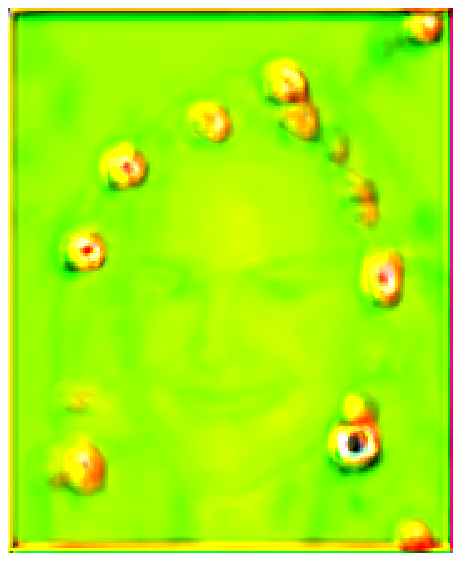}}
    \subfigure[$\lambda=0.5$]{\includegraphics[scale=0.3]{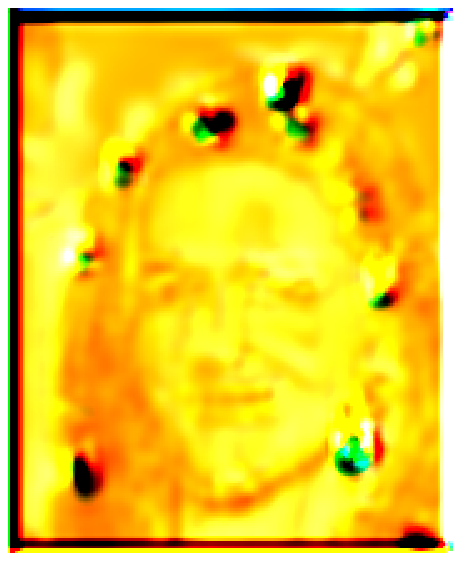}}
    \subfigure[$\lambda=0$]{\includegraphics[scale=0.3]{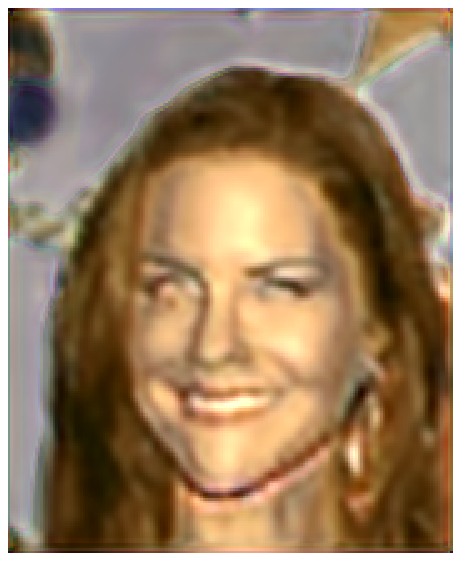}}\vspace{-0.2in}
    \caption{Visualize the impact of the utility-privacy budget $\lambda$ when protecting `gender' in CelebA.}
    \label{fig:visualize}
\end{figure}

We further visualize how different options of $\lambda$ influence the utility maintained by the learned feature extractor. This is done by training a decoder with the reversed architecture of the feature extractor, and then the decoder aims to reconstruct the raw data by taking the extracted feature as input. 
Here we adopt the setting in Figure \ref{fig:budget}(b) as an example, where `gender' is protected when training the feature extractor. 
As Figure \ref{fig:visualize} shows, decreasing $\lambda$ allows a more informative image to be reconstructed. 
This means more information is retained in the extracted feature with a smaller $\lambda$, which is consist with the results shown in Figure \ref{fig:budget}(c).

Additionally, if we compare Figure \ref{fig:budget}(c) vs. Figure \ref{fig:budget}(a-b) and Figure \ref{fig:budget}(f) vs. Figure \ref{fig:budget}(d-e), it can be observed that protecting more private attributes leads to slight degradation in utility with slightly enhanced privacy protection under a particular $\lambda$. 
For example, given $\lambda=0.9$, the accuracy of `smiling' slightly decreases from 90.13\% in Figure \ref{fig:budget}(a) and 89.33\% in Figure \ref{fig:budget}(c) to 88.16\%. 
The accuracy of `young' slightly decreases from 65.77\% in Figure \ref{fig:budget}(a) to 64.85\% in Figure \ref{fig:budget}(c). The reason is that the feature related to the private attributes has some intrinsic correlations to the feature related to the primary learning tasks. 
Therefore, more correlated features may be hidden if more  private attributes need to be protected.

\textbf{Effectiveness of privacy protection:} We quantitatively evaluate the effectiveness of privacy protection offered by TIPRDC by simulating an adversary to infer the private attribute through training a classifier. 
As presented in Table \ref{tb:model_arch_img}, we adopt a default architecture for simulating the adversary's classifier. However, an adversary may train the classifier with different architectures. We implement three additional classifiers as an adversary in our experiments.
The architectural configurations of those classifiers are presented in Table \ref{tb:ablation}. 
We train those classifiers on CelebA by considering recognizing `smiling' as the primary learning task, and `gender' as the private attribute that needs to be protected. 
Table \ref{tab:bruteforce} presents the average classification accuracy for adversary classifiers on testing data. The results show that although we apply a default architecture for simulating the adversary classifier when training the feature extractor, the trained feature extractor can effectively defend against privacy leakage no matter what kinds of architecture are adopted by an adversary in the classifier design.

\begin{table}[t]
\caption{The ablation study with different architecture configurations of the adversary classifier.}\vspace{-0.15in}
\centering
     \begin{tabular}{ccc}
    \hline
    \multicolumn{1}{c|}{V-CL\#16}                                                             & \multicolumn{1}{c|}{V-CL\#19}                                                             & Res-CL                                                                                                                                  \\ \hline
    \multicolumn{3}{c}{Input Feature Maps ($54\times44\times128$)}                                                                                                                                                                                                                                                                  \\ \hline
    \multicolumn{1}{c|}{\begin{tabular}[c]{@{}c@{}}3$\times$conv3-256\\ maxpool\end{tabular}} & \multicolumn{1}{c|}{\begin{tabular}[c]{@{}c@{}}4$\times$conv3-256\\ maxpool\end{tabular}} & \begin{tabular}[c]{@{}c@{}}$\begin{bmatrix}3 \times 3(2), & 128 \\ 3 \times 3,& 128 \end{bmatrix}$\\ $\begin{bmatrix}3 \times 3, & 128 \\ 3 \times 3,& 128 \end{bmatrix}$\end{tabular}   \\ \hline
    \multicolumn{1}{c|}{\begin{tabular}[c]{@{}c@{}}3$\times$conv3-512\\ maxpool\end{tabular}} & \multicolumn{1}{c|}{\begin{tabular}[c]{@{}c@{}}4$\times$conv3-512\\ maxpool\end{tabular}} & $\begin{bmatrix}3 \times 3, & 256 \\ 3 \times 3,& 256 \end{bmatrix} \times 2$                                                           \\ \hline
    \multicolumn{1}{c|}{\begin{tabular}[c]{@{}c@{}}3$\times$conv3-512\\ maxpool\end{tabular}} & \multicolumn{1}{c|}{\begin{tabular}[c]{@{}c@{}}4$\times$conv3-512\\ maxpool\end{tabular}} & $\begin{bmatrix}3 \times 3, & 512 \\ 3 \times 3,& 512 \end{bmatrix} \times 2$                                                           \\ \hline
    \multicolumn{2}{c|}{\begin{tabular}[c]{@{}c@{}}2$\times$fc-4096\\ fc-label length\end{tabular}}                                                                                       & \begin{tabular}[c]{@{}c@{}}avgpool\\ fc-label length\end{tabular}                                                                       \\ \hline
    \end{tabular}\vspace{-2mm}
\label{tb:ablation}
\end{table}

\begin{table}[t]
    \centering
    \caption{The classification accuracy of `gender' on CelebA with different adversary classifiers ($\lambda=0.9$).}\vspace{-0.15in}
    \resizebox{.48\textwidth}{!}{
    \begin{tabular}{|c|c|c|c|c|}
    \hline
        \multirow{2}{*}{\textbf{Training Classifier}} &  \multicolumn{4}{|c|}{\textbf{Adversary Classifier}}\\
        \cline{2-5}
         & Default architecture in Table 1 & V-CL\#16 & V-CL\#19 & Res-CL   \\
        \hline
        Default architecture in Table 1 & 59.02\% & 59.83\% & 60.77\% & 61.05\% \\
        \hline
    \end{tabular}}
    \label{tab:bruteforce}
\end{table}

\textbf{Evaluate the transferability of TIPRDC:} The data collector usually trains the feature extractor of TIPRDC before collecting the data from users. Hence, the transferability of the feature extractor determines the usability of TIPRDC. We evaluate the transferability of TIPRDC by performing cross-dataset evaluations.  Specifically, we train the feature extractor of TIPRDC using either CelebA or LFW dataset and test the utility-privacy tradeoff on the other dataset. In this experiment, we choose recognizing `black hair' as the primary learning task, and `gender' as the private attribute that needs to be protected. As Table \ref{tab:cross} illustrates, the feature extractor that is trained using one dataset can still effectively defend against private attribute leakage on the other dataset, while maintaining the classification accuracy of the primary learning task. For example, if we train the feature extractor using CelebA and then test it on LFW, the accuracy of `gender' decreases to 56.87\% compared with 57.31\% by directly training the feature extractor using LFW. The accuracy of `black hair' marginally increases to 89.27\% from 88.12\%. The reason is that CelebA offers a larger number of training data so that the feature extractor can be trained for a better performance. Although there is a marginal performance drop, the feature extractor that is trained using LFW still works well on CelebA. The cross-dataset evaluations demonstrate good transferability of TIPRDC. \vspace{-0.1in}

\begin{table}[t]
    \centering
    \caption{Evaluate the transferability of TIPRDC with cross-dataset experiments ($\lambda=0.9$).}\vspace{-0.15in}

    \begin{tabular}{c|c|c|c}
    \hline
        \textbf{Training Dataset} & \textbf{Test Dataset} & \textbf{`gender'} & \textbf{`black hair'} \\
        \hline
        LFW & CelebA & 59.73\% & 87.15\% \\
        LFW & LFW & 57.31\% & 88.12\% \\
        \hline
        CelebA & LFW & 56.87\% & 89.27\% \\
        CelebA & CelebA & 58.82\% & 88.98\% \\
        \hline
    \end{tabular}
    \label{tab:cross}
\end{table}

\begin{figure}[t]
    \centering
    \includegraphics[scale=0.3]{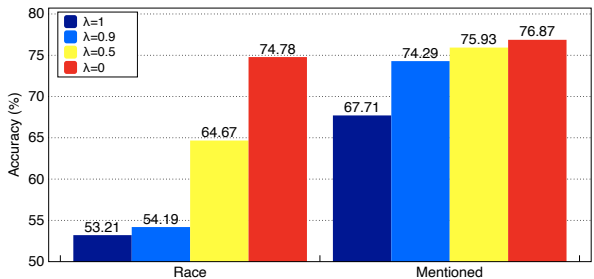}\vspace{-0.15in}
    \caption{The impact of the utility-privacy budget $\lambda$ on DIAL.}
    \label{fig:dial}
\end{figure}

\subsection{Evaluation on DIAL}
To quantitatively evaluate the utility-privacy tradeoff of TIPRDC on DIAl, we choose `race' as the private attribute that needs to be protected and predicting `mentioned' as the primary learning task. The binary mention task is to determine if a tweet mentions another user, i.e., classifying conversational vs. non-conversational tweets. Similar to the experiment settings in CelebA and LFW, we evaluate the utility-privacy tradeoff on DIAL by setting different $\lambda$ with four discrete choices of $\lambda\in\{1, 0.9, 0.5, 0\}$.

As Figure \ref{fig:dial} shows, the classification accuracy of primary learning tasks will increase with a smaller $\lambda$, but the privacy protection will be weakened, showing the same phenomenon as the evaluations on CelebA and LFW. For example, the classification accuracy of `mentioned' is 67.71\% with $\lambda=1$ and increases to 76.87\% by setting $\lambda=0$, and the classification accuracy of `race' increases by 21.57\% after changing $\lambda$ from 1 to 0.

\section{Conclusion}\label{sec:conclusion}
We proposed a task-independent privacy-respecting data crowdsourcing framework TIPRDC.
A feature extractor is learned to hide privacy information features and maximally retain original information from the raw data. By applying TIPRDC, a user can locally extract features from the raw data using the learned feature extractor, and the data collector will acquire the extracted features only to train a DNN model for the primary learning tasks. Evaluations on three benchmark datasets show that TIPRDC attains a  better privacy-utility tradeoff than existing solutions. The cross-dataset evaluations on CelebA and LFW shows the transferability of TIPRDC, indicating the practicability of proposed framework.\vspace{-1mm}

\begin{acks}
This work was supported in part by NSF-1822085 and NSF IUCRC for ASIC membership from Ergomotion. Any opinions, findings, conclusions or recommendations expressed in this material are those of the authors and do not necessarily reflect the views of NSF and their contractors. 
\end{acks}

\bibliographystyle{ACM-Reference-Format}
\bibliography{acmart}

\end{document}